\documentclass[lettersize,journal]{IEEEtran}
\usepackage{hyperref}       
\hypersetup{
    colorlinks=true,
    linkcolor=blue,
    filecolor=magenta,      
    urlcolor=magenta,
    citecolor=blue,
}
\usepackage{bbm}
\usepackage{lipsum}
\usepackage{amssymb}
\usepackage{amsmath}
\usepackage{tabularx}
\usepackage{booktabs}
\usepackage{color, colortbl}
\usepackage{graphicx}
\usepackage{caption}
\usepackage{multirow}
\usepackage{subcaption}
\usepackage{xspace}
\usepackage{lineno}
\newcommand*{\ESCOUTERpositive}{\mbox{E-SCOUTER$^{+}$}\xspace}
\newcommand*{\ESCOUTERnegative}{\mbox{E-SCOUTER$^{-}$}\xspace}
\newcommand*{\ESCOUTER}{\mbox{E-SCOUTER}\xspace}

\DeclareMathOperator*{\argmin}{arg\,min}

\begin{document}

\title{Explainable Image Recognition \\ via Enhanced Slot-attention Based Classifier}

\author{Bowen~Wang,~\IEEEmembership{Member,~IEEE,}
        Liangzhi~Li,~\IEEEmembership{Member,~IEEE,}
        Jiahao~Zhang,
        Yuta~Nakashima,~\IEEEmembership{Member,~IEEE,}
        and~Hajime~Nagahara,~\IEEEmembership{Member,~IEEE,}

\thanks{This work was supported by World Premier International Research Center Initiative (WPI), MEXT, Japan. This work is also supported by JSPS KAKENHI 24K20795 and JST FOREST Grant No. JPMJFR216O. }
\thanks{Authors are with Institute of Datability Science (IDS), Osaka University, 2-8 Yamadaoka, Suita, Osaka 565-0871. (e-mail: wang@ids.osaka-u.ac.jp, liliangzhi@ieee.org, jiahao@is.ids.osaka-u.ac.jp, n-yuta@ids.osaka-u.ac.jp, nagahara@ids.osaka-u.ac.jp)}
\thanks{Bowen Wang, Yuta Nakashima, and Hajime Nagahara are also with Premium Research Institute for Human Metaverse Medicine (WPI-PRIMe), Osaka University, 2-2 Yamadaoka, Suita, Osaka 565-0871. (e-mail: bowen.wang.prime@osaka-u.ac.jp) }
\thanks{\textit{Corresponding author}: Liangzhi Li}
\thanks{Our code is available at https://github.com/wbw520/scouter}
}

\markboth{Journal of \LaTeX\ Class Files,~Vol.~14, No.~8, August~2024}%
{Shell \MakeLowercase{\textit{et al.}}: A Sample Article Using IEEEtran.cls for IEEE Journals}


\maketitle

\begin{abstract}
The imperative to comprehend the behaviors of deep learning models is of utmost importance. In this realm, Explainable Artificial Intelligence (XAI) has emerged as a promising avenue, garnering increasing interest in recent years. Despite this, most existing methods primarily depend on gradients or input perturbation, which often fails to embed explanations directly within the model's decision-making process. Addressing this gap, we introduce \ESCOUTER, a visually explainable classifier based on the modified slot attention mechanism. \ESCOUTER distinguishes itself by not only delivering high classification accuracy but also offering more transparent insights into the reasoning behind its decisions. It differs from prior approaches in two significant aspects: (a) \ESCOUTER incorporates explanations into the final confidence scores for each category, providing a more intuitive interpretation, and (b) it offers positive or negative explanations for all categories, elucidating ``why an image belongs to a certain category" or ``why it does not." A novel loss function specifically for \ESCOUTER is designed to fine-tune the model's behavior, enabling it to toggle between positive and negative explanations. Moreover, an area loss is also designed to adjust the size of the explanatory regions for a more precise explanation. Our method, rigorously tested across various datasets and XAI metrics, outperformed previous state-of-the-art methods, solidifying its effectiveness as an explanatory tool.
\end{abstract}


\section{Introduction}\label{sec:introduction}
\IEEEPARstart{U}{nderstanding} the decision-making process of deep learning models has assumed paramount importance, particularly in domains such as medical diagnosis \cite{amann2020explainability,van2022explainable,reddy2022explainability}, where the consequences of relying on black-box models can be profound. Explainable artificial intelligence (XAI) has been proposed to 
unveil the inner workings of models, uncovering the factors and features influencing their predictions and enabling users to comprehend how and why specific decisions are reached.

The most popular way for XAI to interpret a model's decision involves assessing the importance of pixels or regions within an input image. This has been extensively explored to offer explanation for ``why an image $x$ is classified as category $l$'' \cite{simonyan2013deep,selvaraju2017grad,petsiuk2018rise}. Meanwhile, the notion aiming to clarify ``why it is not classified as category $l$'' \cite{wang2020scout,li2021scouter}, presents an untapped potential for enhancing understanding of the differences between categories. Regardless of the explanation type, one important question that arises here is: \textbf{How do these regions contribute to the decision?}

To explain this, let us consider a fully-connected (FC) classifier for category $l$, represented by $y_l(x)=w^\top_l x+b_l$, where $x$ , $w_l$ , and $b_l$ represent the input image (or its feature vector), learnable vector for $l$, and a learnable scalar, respectively. The training involves identifying $w_l$, which may be decomposed into a set $\mathcal{S}_l = \{s_{li}\}_i$ of discriminative supports, from the training samples, with each support being associated with a specific weight $\gamma_{li}$. The classifier can be rewritten as $y_l(x) = (\sum_i \gamma_{li} s^\top_{li})x + b_l$, where one can always choose $s_{li}$ such that $s_{li}^\top x \geq 0$ by appropriately choosing $\gamma_i$. With this definition, $\gamma_{li} > 0$ means the corresponding $s_{il}$ is a \textit{positive support} for category $l$ that increases $y_l$. Conversely, a $s_{il}$ with $\gamma_{li} < 0$ is a \textit{negative support}. $\mathcal{S}_l$ thus encompasses sets $\mathcal{S}^+_l$ and $\mathcal{S}^-_l$ of positive and negative supports, respectively.

\begin{figure}[t]
\centering
\includegraphics[width=1\columnwidth]{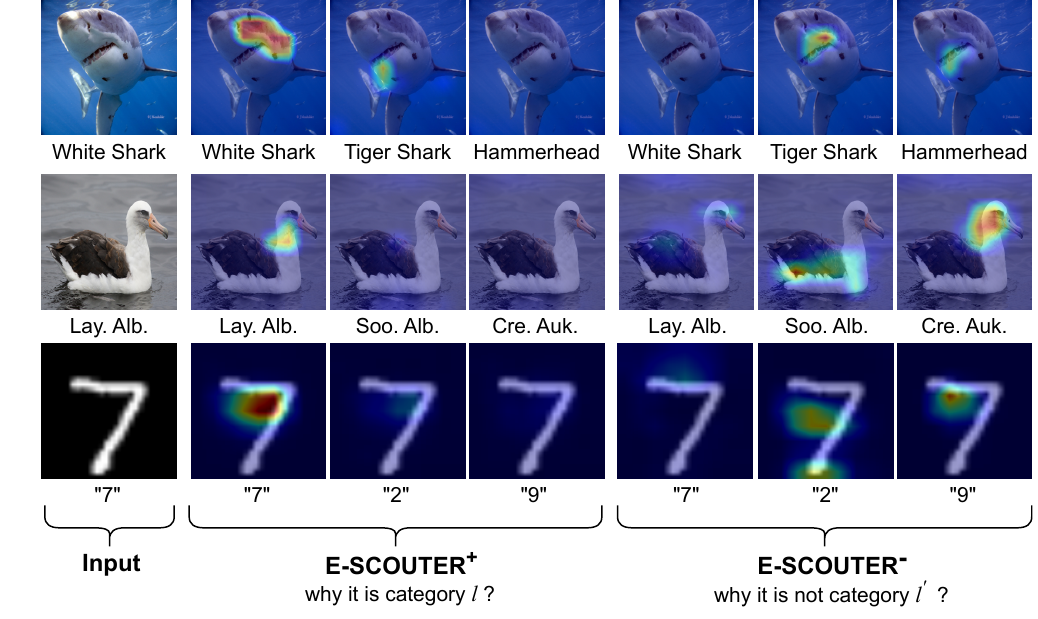}
\caption{Explanations from \ESCOUTER. Using positive ($+$) and negative ($-$) \ESCOUTER losses can emphasize the positive and negative supports respectively, based on which one can understand why or why not the images are classified into a certain category.}
\label{fig1}
\end{figure}

The bottom row of Fig.~\ref{fig1} displays an example from MNIST \cite{MNIST}. The acute angle created by white line segments near the top-right corner can be a positive support for category \texttt{7} as such a structure hardly appears in the other digits. Conversely, the presence of a straight horizontal line segment around the top and the absence of a horizontal line segment around the bottom are negative supports for category \texttt{2}. The former is also a negative support for \texttt{9}. Interpretive supports can also be found in natural image classification tasks, e.g., over shark images (from ImageNet \cite{ImageNet}) and bird images (from CUB200 \cite{CUB-200}). A practical application in medical image analysis, as discussed in \cite{chang2021explaining}, underscores the significance of identifying both positive and negative supports. However, recent mainstream methodologies introduced in \cite{selvaraju2017grad,petsiuk2018rise,wang2020score} have not thoroughly explored this aspect.

In this paper, we introduce \underline{\textbf{E}}nhanced \underline{\textbf{S}}lot-based \underline{\textbf{CO}}nfig\underline{\textbf{U}}rable and Transparent classifi\underline{\textbf{ER}} (\ESCOUTER), which substitutes the FC classifier with the slot-attention mechanism \cite{locatello2020object} to enhance transparency and provide both intuitive and accurate explanations. Unlike existing XAI methods \cite{CAM,selvaraju2017grad,chattopadhay2018grad,petsiuk2018rise,IBA}, \ESCOUTER embeds explanations into the forward decision-making pipeline, ensuring that the rationale for a classification decision is directly accessible and interpretable. Our approach does not merely highlight decision-relevant regions but also elucidates how they contribute, offering insights into the final decision through both positive and negative supports, as illustrated in Fig.~\ref{fig1}. 

A specialized loss function is designed to realize this capability. It allows for fine-tuning the model's behavior, switching between finding positive or negative supports. Moreover, \ESCOUTER is distinguished by a bespoke attention area loss, tailored to constrain the model's focus and provide a more precise explanation. Through extensive experimentation, we demonstrate that \ESCOUTER is not just a competent classifier when compared to FC but also provides superior explanatory insights than previous XAI methods. This makes it an indispensable tool for many applications (e.g., medical diagnosis) where understanding both the ``why" and ``why not" behind a model's decision is critical.

Our contributions are summarized as follows: \textbf{1)} We proposed \ESCOUTER, a transparent classifier, reasoning model decision with both positive and negative supports. \textbf{2)} By utilizing an area size loss functions, our method can control the spatial size of the supports to be learned, facilitating a clearer and more distinct explanation. \textbf{3)} State-of-the-art (SOTA) interpretability across several benchmark datasets. Our comprehensive evaluation demonstrates \ESCOUTER's robustness and versatility in various scenarios. \textbf{4)} The case study in the medical field further emphasizes the significance of both types of supports, along with the necessary to control the area size of the explanatory regions. 

This paper is an extension of our conference paper published in ICCV 2021 \cite{li2021scouter}. In this version, \textbf{1)} we overcome the category limitations, restricted to handling fewer than 200 classification categories, by incorporating an additional normalization step. This improvement also allows our method to have a broader application across various scenarios. \textbf{2)} \ESCOUTER shows a more stable training process. Through several datasets and evaluation metrics, it owns consistently higher classification accuracy and superior interpretability than the former version, as well as, state-of-the-art performance than previous XAI methods.

\section{Related Works}
\subsection{Explainable AI}
Explainable AI methods for image recognition are generally designed to elucidate the relevance of individual pixels of an input image to a model's prediction, culminating in generating a heatmap that highlights relevant regions. Such interpretability sheds light on what leads to a particular decision, making it more transparent and accountable. To accomplish this objective, there are primarily two paradigms \cite{ras2022explainable}, \textit{i.e.}, \textit{post-hoc} and \textit{intrinsic}. \textit{Post-hoc} methods generate a heatmap after the model makes the decision \cite{selvaraju2017grad,petsiuk2018rise,wang2020score}. Conversely, \textit{intrinsic} methods embed heatmap generation in the model's forward path, often leveraging attention modules \cite{kim2017interpretable, mascharka2018transparency}.

The \textit{post-hoc} paradigm has been extensively explored, and the most popular type of method is based on back-propagation. Attribution techniques, \textit{e.g.}, saliency \cite{simonyan2013deep} and its variants \cite{simonyan2013deep,zeiler2014visualizing,springenberg2014striving,smilkov2017smoothgrad,sundararajan2017axiomatic} were introduced to compute the gradient of the input image. While saliency analysis can highlight important regions, it tends to focus on low-level visual features, such as color, contrast, and edges, without considering higher-level semantic information. GradCAM \cite{selvaraju2017grad}, a neuron activation-based technique \cite{Naveed}, combines the model's gradients with the activation map \cite{CAM} of internal neurons. It captures the importance of each spatial location in the feature maps relative to a specific class. Multiple variants have appeared in the literature, \textit{e.g.}, GradCAM++ \cite{chattopadhay2018grad}, Smooth-GradCAM++ \cite{omeiza2019smooth}, and Score-CAM \cite{wang2020score}. All of them use a similar idea to GradCAM. Perturbation provides another aspect of \textit{post-hoc}, which does not rely on a model's structure and learned parameters (i.e., model-agnostic). For instance, Extremal Perturbation \cite{fong2019understanding} iteratively modifies the pixel values of an input image while monitoring the model's response to identify the most salient input pixels. This strategy has been recently improved by designing heuristics to control their solution space as in, e.g., RISE \cite{petsiuk2018rise}, IBA \cite{IBA}, and IGOS \cite{IGOS}. Perturbation methods suffer from computation costs. Some recent works \cite{fel2023don} focus on efficiency.

The conceptual genesis for \textit{intrinsic} methods lies in endowing the inherent interpretability to a deep model \cite{Naveed}. Leveraging attention is a promising way to realize this purpose \cite{kim2017interpretable, mascharka2018transparency, xie2019visual, NEURIPS2022_0073cc73}, while previous methods generally take attention module as a part of the backbone model and are not intended to interpret a classification task. There are also some concept-based frameworks \cite{alvarez2018towards,NEURIPS2019_adf7ee2d,koh2020concept,wang2023botcl} involving the concept-level explanation, which are different from the per-pixel XAI methods (mentioned above). 

SCOUTER \cite{li2021scouter} is an intrinsic method designed to reason classification models in a single forward path. However, it suffers from training instability and limitations regarding the number of classes it can handle, which restrict its broader applicability. This paper introduces \ESCOUTER as an extension to address the issues encountered in the original version for better stability and scalability.

\subsection{Discriminant Counterfactual Explanation}
Given an image of category $l$ and a counterfactual category $l'$, counterfactual explanation \cite{verma2020counterfactual,vandenhende2022making,khorram2022cycle,goyal2019counterfactual,kommiya2021towards} try to find region transformation between this image pair, which can change the model prediction from $l$ to $l'$. Among them, Discriminant Counterfactual Explanation (DCE) \cite{wang2020scout} was introduced to elucidate ``why an image $x$ is classified as category $l$ instead of category $l'$'', employing a more efficient approach by harnessing the attributive explanations \cite{simonyan2013deep,smilkov2017smoothgrad,sundararajan2017axiomatic}. However, its counterfactual explanation comes from the combination of two positive explanations from an image pair, which can only be seen as a proxy of a negative explanation. Notably, several \textit{post-hoc} methods can be extended to provide negative explanations by simply negating the gradient of the prediction score. GradCAM \cite{selvaraju2017grad} refers to its negative variant as a negative explanation, indicating regions that have the potential to alter the model's decision. This strategy may not be straightforward because the linear classifier may lead to the visualization failing to emphasize the support regions for the negative explanation. \ESCOUTER directly generates a heatmap to display important regions within the forward path, revealing the negative part of a model's decision. 

\subsection{Slot-attention for Computer Vision}
Self-attention \cite{vaswani2017attention} has been primarily explored in natural language processing \cite{devlin2018bert}. Self-attention layers process elements in an input sequence one by one, aggregating information across the entire input sequence. This attention mechanism has more recently transferred to computer vision as an essential building block of various models, such as the Image Transformer \cite{parmar2018image}, DEtection TRansformer (DETR) \cite{carion2020end}, and Vision Transformer (ViT) \cite{dosovitskiy2020image}. Slot attention \cite{locatello2020object} is motivated by this self-attention mechanism. It has been employed to extract object-centric features from images. More recent research has extended the utility of slot attention to various computer vision tasks, including feature contrastive learning \cite{wen2022self}, diffusion for image generation \cite{jiang2023object}, and segmentation \cite{xu2022groupvit}. In this paper, we delve into the use of the slot attention mechanism for explainable classification.

\section{Method}
\begin{figure}[t]
    \centering
    \begin{subfigure}[b]{.92\columnwidth}
        \includegraphics[width=\columnwidth]{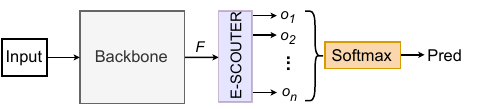}
        \caption{}
        \label{fig:overview1}
    \end{subfigure}
    \begin{subfigure}[b]{1\columnwidth}
        \includegraphics[width=\columnwidth]{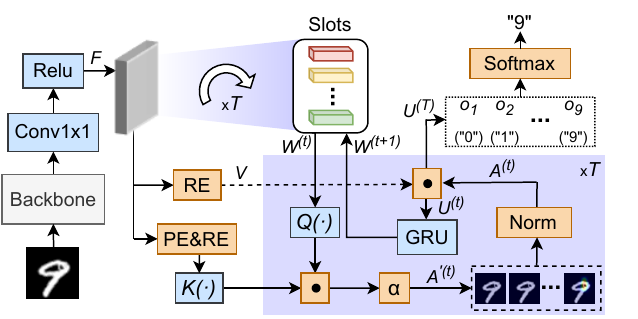}
        \caption{}
        \label{fig:overview2}
    \end{subfigure}
    \caption{Classification pipeline. (a) \ESCOUTER as a classifier. (b) The overview of \ESCOUTER for classification, where PE is position embedding, RE is reshape operation, $\sigma$ is sigmoid activation, and ($\cdot$) denotes dot multiplication.}
    \label{overview}
    \vspace{-0.1in}
\end{figure}

The primary goal of an image classification model given an input image $x$ is to determine the most possible category $l$ it belongs to, from a predefined set $\mathcal{L} = \{l_1, l_2, \ldots, l_n\}$ of categories. This can be done by using a backbone model $B$ to extract feature map $F=B(x) \in \mathbb{R}^{h \times w \times c}$, where $h$, $w$, and $c$ are the height, width, and the number of channels, respectively. After a global average pooling, a multilayer perceptron (MLP) classifier with FC layers and softmax converts $F$ into confidence scores $o = \{o_1, o_2, \ldots, o_n\}$ for respective categories in $\mathcal{L}$. Although FC provides substantial learning capacity, it also makes it difficult to determine which features are crucial for the prediction outcome and how these features interact with each other.

\ESCOUTER substitutes for the MLP classifier to compute $o$ in order for explainable image classification. As illustrated in Fig.~\ref{fig:overview1}, the entire network, including the backbone, is trained with a customized loss function. This loss function enables control over the desired size of relevant regions and offers the switch between finding positive and negative supports.

\subsection{\ESCOUTER}
The key component of \ESCOUTER is a slot-attention module \cite{locatello2020object}, which initially proposes \textit{slots} to represent regions aggregated with the attention mechanism \cite{vaswani2017attention}. Each slot is responsible for exclusively attending to a single visual concept (e.g., a pattern and a part of an object) within the image through the dedicated attention map, thereby producing distinct features of the concept as output. This mechanism can be extended to identify relevant regions to a specific decision within an input image. What follows details \ESCOUTER. 

As illustrated in Figure \ref{fig:overview2}, the initial feature map $F$ undergoes a transformation process. It starts with a $1\times 1$ convolution layer, followed by the activation of a ReLU nonlinearity as $F^* = \operatorname{ReLU}(\operatorname{Conv}(F))$. These transformations reduce the dimensionality of $F$ from $c$ to $d$. Then, the spatial dimension of $F^*$ is flattened to obtain a feature matrix $V \in \mathbb{R_+}^{d \times s}$, where $s = hw$. To preserve spatial information, a positional embedding $\text{PE}$ \cite{vaswani2017attention,locatello2020object,li2021scouter} is added to the features, denoted as $\tilde{F} = F^* + \text{PE}$, whose spatial dimension is also flattened.

\ESCOUTER uses a single slot-attention module with $n$ slots as a classifier, which is applied to feature map $F$ (Fig.~\ref{fig:overview1}). Each slot is associated with a specific category $l \in \mathcal{L}$ and is responsible for identifying a support of $l$ that increases (or decreases) the confidence score for $l$. A slot for $l$ is a learnable vector $w_l$ that describes a visual concept associated with $l$, which can be collectively denoted by $W \in \mathbb{R}^{n \times d}$ for all $l \in \mathcal{L}$.

\if 1
\textcolor{red}{As the reason for using GRU here \cite{},  \ESCOUTER iteratively updates $w_l$ through gated recurrent units (GRU) $T$ times. 
Specifically, $w_l$ is randomly initialized as
\begin{equation}
    w_l \sim \mathcal{N}(\mu, \sigma I) \in \mathbb{R}^{1 \times d},
\end{equation}
where $\mu$ and $\sigma$ represent the mean and variance of a Gaussian distribution, $I$ is the identity matrix, and $d$ denotes the dimensionality of the weight vector. }
\fi

The original slot-attention module updates the slots with a gated recurrent unit (GRU) for more defined responses around the edges between the objects of interest and the rest. We also adopt this design choice in \ESCOUTER. Let $W^{(t)}$ be the slots after the $t$-th iteration, where $t = 1, 2, \ldots, T$ and $W^{(1)} = W$. We employ two multilayer perceptrons (MLPs), denoted as $Q$ and $K$, each comprising three FC layers with ReLU nonlinearities in between. These MLPs transform $W^{(t)}$ and $\tilde{F}$ to be `query' and `key' as
\begin{equation}
    Q(W^{(t)}) \in \mathbb{R}^{n \times d}, \quad K(\tilde{F}) \in \mathbb{R}^{d \times s}.
\end{equation}
\ESCOUTER uses the dot-product similarity \cite{vaswani2017attention} between $W$ and $\tilde{F}$ to compute attention matrix $A^{(t)}$
\begin{equation} \label{eq_QxK}
    A^{(t)} = \sigma(Q(W^{(t)}) K(\tilde{F})) \quad \in (0,1)^{n \times s},
\end{equation}
where $\sigma$ is the element-wise sigmoid. This attention matrix encapsulates the dependency between the slots and spatial elements within the feature map, showing the spatial positions in the feature map at which the concept associated with each slot appears.

We further normalize the attention weights to compactify the region. Let $a^{(t)}_l$ be the row vector for category $l$. We normalize the activation over the spatial positions by
\begin{align}
    \bar{a}^{(t)}_l = \frac{a^{(t)}_l}{a^{(t)}_l \mathbf{1}_s + 1}, \label{eq:normalization}
\end{align}
where $\mathbf{1}_s$ is a column vector in $\mathbb{R}^s$ with all elements being 1. The normalized attention matrix is denoted by $\bar{A}^{(t)}$. The attention weights are then applied to $V$ to obtain the features associated with each slot:
\begin{align} 
    U^{(t)} = \bar{A}^{(t)} {V}^\top \quad \in \mathbb{R}_+^{n \times d}.\label{eq:feature}
\end{align}

The normalization by Eq.~(\ref{eq:normalization}) is a critical step forward from the original SCOUTER \cite{li2021scouter}, effectively mitigating excessive attention values that could impede training progress, while preserving the magnitude of smaller activation. 

The slots $W^{(t)}$ are then updated through a GRU as:
\begin{equation} \label{eq_GRU}
W^{(t+1)}=\operatorname{GRU}(U^{(t)}, W^{(t)}).
\end{equation}
This operation takes $U^{(t)}$ and $W^{(t)}$ as input, giving $W^{(t+1)}$ as the hidden state. Following the original slot-attention, $T$ is defaulted to 3.

The output of \ESCOUTER is the summation of all feature values corresponding to category $l$ in $U^{(T)}$, i.e.,
\begin{equation}\label{eq:cls}
o = U^{(T)} \mathbf{1}_{d} \quad \in \mathbb{R}_+^n.
\end{equation}
Eq.~(\ref{eq:feature}) and Eq.~(\ref{eq:cls}) give a vivid interpretation of \ESCOUTER: 
\begin{equation}\label{eq:merge}
o=\bar{A}^{(T)}V^\top\mathbf{1}_{d} \quad \in \mathbb{R}_+^n.
\end{equation}

It is worth noting that, in contrast to the original slot attention module, \ESCOUTER omits a linear transformation applied to the features $V$, as it already possesses a sufficient number of learnable parameters in the convolution layer $\text{Conv}$ applied to $F$, $Q$, $K$, GRU, and others, ensuring flexibility. 

\noindent\textbf{Interpretation}. The elements of $V$ may be interpreted as the indicators of the presence of a support (or concept) for one of the categories\footnote{It should be noted that each element of the column vectors of $V$ is not necessarily an indicator of a support; some elements in a column vector can combinatorially represent one.}, and $V^\top\mathbf{1}_{d} \in \mathbb{R}^{d}$ can be then seen as a map that shows the presence of supports for \textit{any} category. Meanwhile, $\bar{A}$ gives the spatial positions corresponding to each slot (or category). Therefore, $o$, which is a multiplication of $\bar{A}$ and $V^\top\mathbf{1}_{d} \in \mathbb{R}^{d}$, gives the weighted sum of support indicators for each category. 

Typical classifiers use FC layers to compute the confidence for each category, while \ESCOUTER uses Eq.~(\ref{eq:cls}), which is the weighted sum of the support indicators. An obvious advantage of this design is the interpretability of the decision-making process, as the regions with a larger weighted value have a greater influence on the final prediction. Therefore, by visualizing the attention map, we can directly know which regions are supporting the model's final decision. 

\subsection{\ESCOUTER Loss}
The overall pipeline is implemented in an end-to-end manner and can be trained by minimizing the softmax cross-entropy loss $\ell_\text{CE}$ of confidences $o$ and ground-truth label $y \in \mathcal{L}$. 

With the design of \ESCOUTER, we presume that the supports for a category, or equivalently the regions specified by $\bar{A}$, only occupy smaller areas in the image because of the locality of visual elements, which is encoded in Eq.~(\ref{eq:normalization}). The expected areas, though, can vary for different domains (or datasets). More explicit control over the areas can be advantageous to encode our beliefs about them. Hence, we introduce the area loss $\ell_{\operatorname{Area}}$ to regulate the size of support regions, given by:
\begin{equation}
\ell_{\operatorname{Area}} = \sum_{l \in \mathcal{L}} \mathbf{1}_s^\top \bar{a}_l^
{(T)} \label{eq:area_loss}
\end{equation}
summing all the elements in $\bar{A}^{(T)}$. 

The overall loss is:
\begin{equation} \label{eq:loss}
\ell_{\operatorname{E-SCOUTER}} = \ell_{\operatorname{CE}}+ \lambda \ell_{\operatorname{Area}},
\end{equation}
where $\lambda$ serves as a factor to adjust the weight of the area loss. By increasing the value of $\lambda$, \ESCOUTER tends to focus on smaller regions. Conversely, a smaller value of $\lambda$ encourages the model to prioritize larger areas. The ablation study in Section~\ref{ablation} provides additional insights and details.

\subsection{Positive and Negative Explanation}
The \ESCOUTER loss in Eq.~(\ref{eq:loss}) only offers a positive explanation and we introduce a hyper-parameter $e \in \{+1, -1\}$ to Eq.~(\ref{eq:cls}) as: 
\begin{equation}\label{eq:posneg}
o = e \cdot U^{(T)} \mathbf{1}_{d} \quad \in \mathbb{R}_+^n.
\end{equation}
This hyper-parameter is pivotal in configuring \ESCOUTER to learn either positive or negative supports. Since all elements of $U^{(T)}$ are constrained to be non-negative, the model exhibits distinct behaviors for different values of $e$. 

Utilizing the softmax activation and cross-entropy loss, the model is trained to assign the highest confidence value, $o_l$, to the ground-truth category $l = y$, while giving smaller values, $o_{l}$, to the others (i.e., $l \neq y$). As both $\bar{A}^{(T)}$ and $V$ are non-negative, all elements in $o$ are non-negative as well. 

For $e=+1$, larger values of $o_l$ can only be produced when some elements in $\bar{a}^{(T)}_l$, which is the row vector in $\bar{A}^{(T)}$ corresponding to category $l$, are close to 1. Conversely, smaller values of $o_{l}$ are obtained when all elements in $\bar{a}^{(T)}_{l}$ are close to 0. Therefore, when $e$ is set to $+1$, the model learns to identify \textit{positive} supports $\mathcal{S}^{+}_{l}$ as a larger value of an element in $V$ positively amends $o_l$. The visualization of $\bar{a}^{(T)}_l$ thus serves as a positive explanation, as illustrated in Fig.~\ref{fig1} (left).

In contrast, when $e = -1$, all elements in $o$ are negative, resulting in $o_l$ being close to 0 when $l = y$, while $o_l$ gives a smaller value when $l \neq y$. Due to the non-negativity of $V$, all elements in $\bar{a}^{(T)}_{l}$ must be close to 0 for $l=y$, whereas smaller values of $o_{l}$ are generated when $\bar{a}^{(T)}_{l'}$ includes elements close to 1. The model thus learns to discover \textit{negative} supports $\mathcal{S}^{-}_{l}$, which do not manifest in images of the ground-truth category. Consequently, $\bar{a}^{(T)}_{l'}$ can be used as a negative explanation, as illustrated in Fig.~\ref{fig1} (right).

\section{Experiments}
\subsection{Experimental Settings}
We evaluated \ESCOUTER using two widely-recognized datasets: ImageNet \cite{deng2009imagenet} and CUB200 \cite{CUB-200}. These datasets benchmark the performance of classification models. To cover small-scale tasks, which may be more realistic, we also conducted evaluations on the MNIST \cite{MNIST} and Caltech \cite{caltech} datasets. Additionally, we undertook two case studies in medical image analysis. The first case study focuses on diagnosing glaucoma using the ACRIMA dataset \cite{ACRIMA}, and the second involves multi-label classification for chest x-ray diagnosis using the X-ray14 dataset \cite{wang2017chestx}.

All images used in our evaluation are the size of $260\times 260$. The number of channels $c$ in $F$ is due to the backbone, and in $F'$ is $d=64$. Following the original slot-attention \cite{locatello2020object}, the iterations of GRU defaulted to $T=3$. All models underwent training on the training set of respective datasets with a batch size of 256 for 40 epochs, and performance metrics were computed on the validation set using the trained models at the last epoch. We employed the AdamW optimizer \cite{loshchilov2017decoupled} with an initial learning rate of $10^{-4}$, which was reduced by a factor of ten after 30 epochs. Only horizontal flipping is used as data augmentation. The implementations of the underlying CNN models were sourced from the PyTorch timm library \cite{rw2019timm}. All our experiments were done with a GPU server with 4 NVIDIA A100.

\begin{figure*}[!t]
	\centering
	\includegraphics[width=1\linewidth]{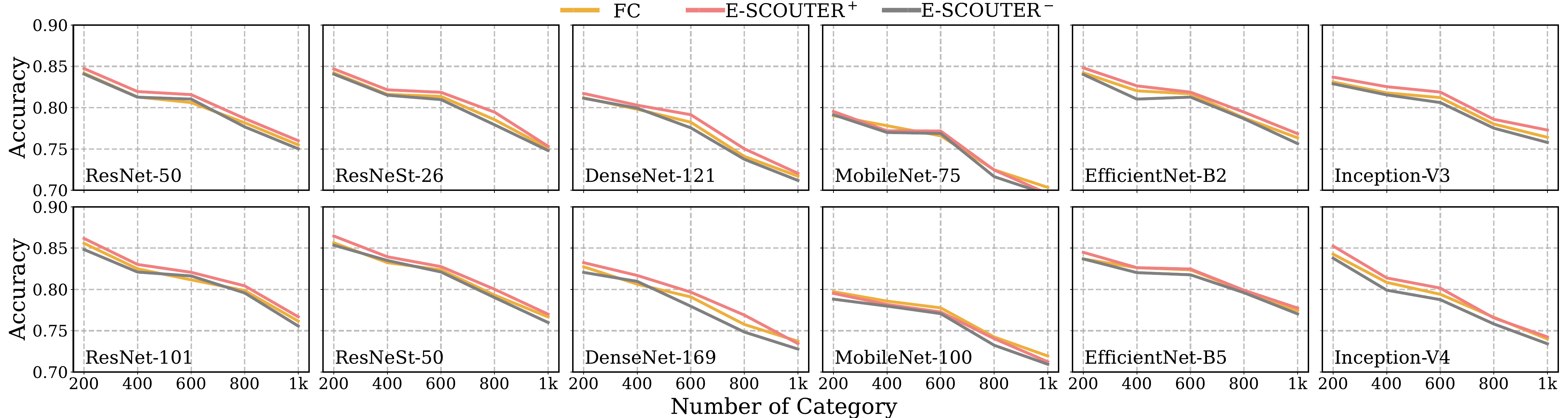}
	\caption{Classification performance of different models with FC classifier, \ESCOUTERpositive ($\lambda=10$), and \ESCOUTERnegative ($\lambda=10$). The horizontal axis is the number of categories, where the first $n$ categories of the ImageNet dataset were used; the vertical axis is the accuracy.}
	\label{acc_vs_cat}
\end{figure*}

\subsection{Evaluation Metrics}\label{sec:metrics}
We utilize six metrics to assess the performance of \ESCOUTER in comparison to other XAI methods. Metrics are computed based on the visualization results. To evaluate positive explanations, the ground-truth object region, provided for each image, covering the region corresponding to the ground-truth label $y$,\footnote{Some datasets provide ground-truth object regions.} is used as their ground-truth region. On the other hand, each image has $n-1$ negative explanations, one for each non-ground-truth category. For an image of category $y$, we identify the least similar category (LSC) $l'$ to $y$ to evaluate the negative explanation. This choice is based on the observation that LSC images consistently exhibit unequivocal negative explanations. 

To find LSC, we use the semantic similarity between categories $y$ and $l \in \mathcal{L}$ based on \cite{Wu_Palmer} as
\begin{equation} \label{eq_wordnet_similarity}
l' = \argmin_{l \in \mathcal{L}} \frac{2\cdot\operatorname{Depth}(\operatorname{LCM}(y,l))}{\operatorname{Depth}(y)+\operatorname{Depth}(l)},
\end{equation}
where $\operatorname{Depth}(\cdot)$ provides the depth of the category in WordNet \cite{miller1995wordnet}, and $\operatorname{LCM}(y,l)$ identifies the lowest common ancestor of categories $y$ and $l'$. For the CUB200 dataset, WordNet is not applicable because there is no accurate relation for bird species, so we computed similarity based on category-wise feature embedding using the following equation:
\begin{equation} \label{eq_cos_similarity}
l' = \argmin_{l \in \mathcal{L}} \cos(E(y), E(l)),
\end{equation}
where $\cos(\cdot)$ denotes cosine similarity, and $E(\cdot)$ represents the average of feature embeddings of a category's training samples, extracted from a pre-trained CLIP \cite{radford2021learning}.

It should be noted that, regardless of positive and negative explanations, the object regions used as ground truth should be viewed as just a rough \textit{proxy} of the desired explanation of respective XAI methods. For \ESCOUTER, a positive explanation should pinpoint a visual concept (associated with the slot), which should be at least in the object region but does not necessarily cover the whole region.

We use the following six metrics for evaluation. In the definitions, we use $x$ and $\bar{x}$ to represent the input image and the object region for the target category (i.e., $y$ for a positive explanation and $l'$ for a negative explanation), as well as confidence $c(x) \in \mathcal{R}$. The object region can be derived from the bounding box sourced from ImageNet \cite{deng2009imagenet} or the segment mask in CUB200 \cite{CUB-200}. We also denote the relevance map for the target category as $r(x)$, which is resized to match $x$'s size.\footnote{For positive \ESCOUTER, $c(x) = o_y$ and $r = a^{(T)}_{y}$.} 

(i) \textbf{Precision} represents a broadened version of the pointing game \cite{top_down}. As higher relevance scores should be concentrated in $\bar{x}$ for a successful explanation, Precision measures the degree of the concentration by 
\begin{align}
    \text{Precision} = \frac{\sum_{p \in \bar{x}} r_p(x)}{\sum_{p \in x} r_p(x)},
\end{align} 
where $r_p(x)$ give the relevance score of pixel $p$.

(ii) \textbf{Insertion} \cite{petsiuk2018rise} is determined by incrementally adding pixels, prioritized by their relevance, to an initially blank image and observing the evolution of confidence. A successful explanation should give a higher relevance score to a pixel relevant to the category, resulting in a swift increase in confidence along with adding more pixels. Insertion is defined as the area under the curve of the number of added pixels versus the confidence. 

(iii) \textbf{Deletion}\cite{petsiuk2018rise} is computed by gradually removing pixels from the original image based on their relevance, in contrast to the insertion. Confidence should rapidly decline if the explanation is meaningful. Deletion is defined in the same way as the insertion.

(iv) \textbf{Infidelity} \cite{yeh2019fidelity} measures how much perturbation $\phi \in \mathbb{R}^{\gamma}$ to the input image $x \in \mathbb{R}^{\gamma \times 3}$ impacts the confidence and relevance map, given by
\begin{equation}\label{f2_spp}
    \text{Infidelity} = \mathbb{E}_{\phi}[(\phi^{\top}r-(c(x)-c(x-\phi)))^2],
\end{equation}
where $\gamma$ is the number of pixels in the input image, and the expectation is computed for $\phi \sim \mathcal{N} (0,\sigma^2)$ with $\sigma = 0.2$.\footnote{In Eq.~(\ref{f2_spp}), we slightly abuse the notation $x-\phi$ to represent subtraction with broadcasting $\phi$ to all channels of $x$.}

(v) \textbf{Stability} \cite{alvarez2018towards} gauges the stability of an XAI method when the input undergoes minor perturbations, such as Gaussian noise, given by
\begin{equation}\label{f1_spp}
    \operatorname{Stability} = \frac{\| r(x) - r(x') \|_2}{\| x - x' \|_2},
\end{equation}
where $x'$ is $x$ with minor white noise. Adding minor noise hardly change the confidence, but the relevance map may change a lot if the method is unstable. For this metric, the smaller the better.

(vi) \textbf{Time} required to obtain the explanation, including feature extraction for classification, is also measured as a metric for computational burden. 

For metrics (ii)-(v), we use the official implementation released in GitHub. For metric (vi), we measure the average time over 10,000 samples from ImageNet's validation set.

\subsection{\ESCOUTER as a Classifier} \label{classifier}
\ESCOUTER serves as an alternative to FC classifiers, offering intrinsic interpretability. It is essential to quantify its classification efficacy across a variety of backbone models and datasets. Fig.~\ref{acc_vs_cat} illustrates the performance of both positive and negative variants of \ESCOUTER (denoted as \ESCOUTERpositive and \ESCOUTERnegative, respectively) on ImageNet. The horizontal and vertical axes are the number of first $n$ categories of the original ImageNet categories used during training and the accuracy. An FC classifier, \ESCOUTERpositive, and \ESCOUTERnegative are represented by yellow, red, and grey lines, respectively. As the category number increases, the performance declines for the FC classifier and \ESCOUTER{}s. 

\begin{table}[!t]
    \caption{Classification accuracy across various datasets using $\lambda = 10$ and ResNet \cite{he2016deep} variants as the backbone.}
    \label{table_exp_classification_other_datasets}
    \centering
    \resizebox{1\columnwidth}{!}{%
        \begin{tabular}{llccc}
            \toprule
            Models & Classifier  &  ImageNet \cite{deng2009imagenet} & Caltech \cite{caltech}  & CUB200 \cite{CUB-200}\\
            \midrule
            \multirow{3}{*}{ResNet-18} 
            & FC & 0.6606 & 0.7918 & 0.7325\\
            & \ESCOUTER$^+$ & \underline{\textbf{0.6675}} & \underline{\textbf{0.8058}} & \underline{\textbf{0.7240}} \\
            & \ESCOUTER$^-$ & 0.6589 & 0.7950 & 0.7056 \\
            \midrule
            \multirow{3}{*}{ResNet-50}
            & FC & 0.7550 & 0.8470 & \underline{\textbf{0.7850}} \\
            & \ESCOUTER$^+$ & \underline{\textbf{0.7604}} & \underline{\textbf{0.8615}} & 0.7730 \\
            & \ESCOUTER$^-$ & 0.7522 & 0.8452 & 0.7544 \\
            \midrule
            \multirow{3}{*}{ResNet-101}
            & FC & 0.7616 & 0.8755 & \underline{\textbf{0.7984}} \\
            & \ESCOUTER$^+$ & \underline{\textbf{0.7638}} & \underline{\textbf{0.8794}} & 0.7901 \\
            & \ESCOUTER$^-$ & 0.7572 & 0.8687 & 0.7763 \\
            \bottomrule
        \end{tabular}
    }
\end{table}

\begin{table}[t]
\caption{Comparison of \ESCOUTER and an FC classifier on ImageNet \cite{deng2009imagenet} in terms of required computational resource ($n$ is 1,000 and the size of input images are $260\times260$).}
\centering
\resizebox{1\columnwidth}{!}{
\begin{tabular}{lcccc}
\toprule
&\multicolumn{2}{c}{Params (M)} &\multicolumn{2}{c}{Flops (G)} \\
\cmidrule(lr){2-3} \cmidrule(lr){4-5} 
Model & FC  & \ESCOUTER & FC & \ESCOUTER \\ 
\midrule
ResNet-18 \cite{he2016deep} & 11.6895 & \underline{\textbf{11.3107}} & \underline{\textbf{2.6527}} & 2.7321  \\ 
ResNet-50 \cite{he2016deep} & 25.5570 & \underline{\textbf{23.7406}} & \underline{\textbf{6.0158}} & 6.1016\\
ResNeSt-26 \cite{zhang2022resnest} & 17.0694 & \underline{\textbf{15.2530}} & \underline{\textbf{5.2087}} & 5.2945 \\
ResNeSt-50 \cite{zhang2022resnest} & 27.4832 & \underline{\textbf{25.6668}} & \underline{\textbf{7.7841}} & 7.8699 \\
\midrule
DenseNet-121 \cite{huang2017densely} & 7.9788 & \underline{\textbf{7.1208}} & \underline{\textbf{3.7535}} & 3.8337 \\
DenseNet-169 \cite{huang2017densely} & 14.1494 & \underline{\textbf{12.6924}} & \underline{\textbf{4.4399}} & 4.5221 \\
MobileNet-75 \cite{howard2017mobilenets} & 2.5106 & \underline{\textbf{1.9372}} & \underline{\textbf{0.2447}} & 0.3249  \\
MobileNet-100 \cite{howard2017mobilenets} & 3.9329 & \underline{\textbf{3.1348}} & \underline{\textbf{0.3409}}  & 0.4221  \\
\midrule
EfficeintNet-B2 \cite{tan2019efficientnet} & 9.1099 & \underline{\textbf{7.8926}} & \underline{\textbf{1.0541}} & 1.1372 \\
EfficeintNet-B5 \cite{tan2019efficientnet} & 30.3897 & \underline{\textbf{28.5733}} & \underline{\textbf{3.7067}} & 3.7925  \\
Inception-V3 \cite{szegedy2016rethinking} & 23.8345 & \underline{\textbf{22.0181}} & \underline{\textbf{3.9594}} & 4.0388  \\
Inception-V4 \cite{szegedy2017inception} & 42.6798 & \underline{\textbf{41.3426}} & \underline{\textbf{8.5120}} & 8.5906  \\
\bottomrule
\end{tabular}}
\label{efficeincy}
\end{table}

\begin{table*}[t]
\caption{Evaluation of the explanations based on the ResNet-50 backbone over ImageNet \cite{deng2009imagenet}  and CUB200 \cite{CUB-200} in terms of Precision (Pre.), Insertion (Ins.), Deletion (Del.), Infidelity (Inf.) and Stability (Sta.). Time (in second) is measured with 10,000 samples from the ImageNet validation set on a single NVIDIA A100 GPU.}
\centering
\resizebox{1\linewidth}{!}{
\begin{tabular}{clccccccccccc}
\toprule
 & &\multicolumn{5}{c}{CUB200 \cite{CUB-200}} &\multicolumn{6}{c}{ImageNet\cite{ImageNet}}\\
\cmidrule(lr){3-7} \cmidrule(lr){8-13} 
& Methods & Pre. $\uparrow$ & Ins. $\uparrow$ & Del. $\downarrow$ & Inf. $\downarrow$ & Sta. $\downarrow$ & Pre. $\uparrow$ & Ins. $\uparrow$ & Del. $\downarrow$ & Inf. $\downarrow$ & Sta. $\downarrow$ & Time (s)\\
\midrule
\multirow{15}{*}{Positive} & Saliency \cite{simonyan2013deep}  & .4012 & .5562 & .1299 & .6008 & .1078 & .5644 & .5701 & .1417 & .8129 & .0968 & .0470\\
& SmoothGrad \cite{smilkov2017smoothgrad}  & .6408 & .6819 & .1553 & .6107 & \underline{\textbf{.1006}} & .6137 & .6537 & .1495 & .6834 & .0743 & .8950\\
& InteGrad \cite{sundararajan2017axiomatic}  & .6217 & .6728 & .1599 & .6178 & .1106 & .5987 & .6502 & .1545 & .6937 & .0817 & .8420\\
& Occlusion \cite{zeiler2014visualizing}  & .4145 & .5850 & .1566 & .6623 & .2483 & .5472 & .5723 & .2352 & .7751 & .1761 & .2210\\
& GradCAM \cite{selvaraju2017grad}  & .6301 & .7474 & .0615 & .6428 & .1462 & .6850 & .6840 & .1308 & .8661 & .0714 & \underline{\textbf{.0140}}\\
& GradCAM++ \cite{chattopadhay2018grad}  & .6355 & .7535 & .0628 & .6470 & .1323 & .7097 & .6921 & .1270 & .7908 & \underline{\textbf{.0613}} & .0240\\
& ScoreCAM \cite{wang2020score}  & .6517 & \underline{\textbf{.7601}} & .0594 & .6271 & .1242 & .7128 & .7003 & .1241 & .7728 & .0627 & .1558\\
& GroupCAM \cite{zhang2021group}  & .6449 & .7580 & .0688 & .6140 & .1208 & .7089 & \underline{\textbf{.7115}} & .1327 & .7583 & .0644 & .1380\\
& SESS \cite{tursun2022sess}  & .6625 & .7517 & .0651 & .6207 & .1524 & .7020 & .7036 & .1356 & .7152 & .1024 & .0512\\
& RISE \cite{petsiuk2018rise}   & .5370 & .7312 & .0603 & .6039 & .2377 & .5564 & .6875 & .1315 & .5590 & .2541 & 5.179 \\
& Extremal Perturbation \cite{fong2019understanding}  & .5012 & .7250 & .1108 & .6361 & .2542 & .5297 & .6383 & .1555 & .5821 & .2957 & 4.826 \\
& IBA \cite{IBA}   & .5715 & .7084 & .1024 & .6159 & .1771 & .6440 & .7027 & .1363 & .6411 & .1554 & .8960 \\
& IGOS++ \cite{khorram2021igos++}  & .5543 & .6018 & .1564 & .7155 & .3472 & .6969 & .6081 & .2265 & .6137 & .2773 & 6.027\\
& Greedy-AS \cite{GreedyAS}  & .5960 & .6817 & .1359 & .6380 & .1854 & .6101 & .6152 & .1715 & .6440 & .1513 & 7915. \\
\cmidrule(lr){2-13} 
& \ESCOUTER$^+$$_{\lambda=1}$  & .6838 & .7329 & .0537 & \underline{\textbf{.4682}} & .2497 & .7200 & .7030 & .1367 & \underline{\textbf{.3982}} & .1853 & .2200\\
& \ESCOUTER$^+$$_{\lambda=3}$  & .6910 & .7301 & \underline{\textbf{.0495}} & .5087 & .2202 & .7391  & .6980 & .1225 & .4207 & .1138 & .2160\\
& \ESCOUTER$^+$$_{\lambda=10}$  & \underline{\textbf{.7029}} & .7183 & .0514 & .5133 & .2053 & \underline{\textbf{.7459}} & .6802 & \underline{\textbf{.1184}} & .4412 & .0987 & .2090\\
\midrule
\multirow{8}{*}{Negative} & Saliency \cite{simonyan2013deep}  & .5173 & .6673 & .0842 & .6535 & .1212 & .5437 & .5298 & .1644 & .6075 & .1396 & .0420\\
& SmoothGrad \cite{smilkov2017smoothgrad}  & .6350 & .6742 & .1018 & .6498 & \underline{\textbf{.0798}} & .6081 & .6125 & .1710 & .6351 & .1051 & .8890\\
& Occlusion \cite{zeiler2014visualizing}  & .4749 & .5398 & .2541 & .7278 & .2005 & .5240 & .5571 & .2114 & .7508 & .1737 & .2230\\
& GradCAM \cite{selvaraju2017grad}  & .6881 & .6965 & .0721 & .6597 & .1078 & .6469 & .6391 & .1485 & .8209 & .0621 & \underline{\textbf{.0151}} \\
& GradCAM++ \cite{chattopadhay2018grad}  & .6920 & \underline{\textbf{.6990}} & .0702 & .6517 & .1086 & .6563 & \underline{\textbf{.6475}} & .1438 & .8071 & \underline{\textbf{.0619}} & .0263\\
\cmidrule(lr){2-13} 
& \ESCOUTER$^-$$_{\lambda=1}$  & .7136 & .6496 & .0608 & .5988 & .1130 & .7103 & .6028 & .1381 & .3755 & .1046 & .2236\\
& \ESCOUTER$^-$$_{\lambda=3}$  & .7180 & .6717 & \underline{\textbf{.0575}} & \underline{\textbf{.5839}} & .1235 & .7241 & .6220 & \underline{\textbf{.1306}} & \underline{\textbf{.3468}} & .0876 & .2184\\
& \ESCOUTER$^-$$_{\lambda=10}$  & \underline{\textbf{.7285}} & .6609 & .0580 & .6117 & .1199 & \underline{\textbf{.7326}} & .6107 & .1450 & .4514 & .0752 & .2170\\
\bottomrule
\end{tabular}
}
\label{acc_tab}
\end{table*}

Notably, \ESCOUTERpositive consistently outperforms FC classifiers in terms of classification accuracy across different backbone models. \ESCOUTERnegative, while still comparable to the FC classifiers for a smaller number of categories, exhibits a slight decrease by approximately less than 1\% in some backbones, like DenseNet \cite{huang2017densely}, when $n$ comes closer to 1000. This observation suggests that \ESCOUTER is competent even for large datasets, addressing a major limitation of the original SCOUTER \cite{li2021scouter}. The improvement over the original is elaborated in Section \ref{compare_previous}.

We further extended our experiments to the CUB200 \cite{CUB-200} and Caltech \cite{caltech} datasets using the ResNet variants \cite{he2016deep} as a backbone (Table \ref{table_exp_classification_other_datasets}). The results demonstrate that \ESCOUTER generalizes well across different domains and achieves performance comparable to the FC classifiers. However, a notable limitation is observed in its application to CUB200, particularly with \ESCOUTERnegative. Unlike ImageNet, CUB200 is designed to differentiate among similar bird images, and each category contains a limited number of images (50 per class). These characteristics make it challenging to identify consistent and effective supports that appear across all images within a category.

Table \ref{efficeincy} presents a computational cost of \ESCOUTER{}s and the FC classifiers. \ESCOUTER incurs floating-point operations per second (FLOPS) that are slightly higher than that of the FC classifiers, while it maintains a slightly lower number of parameters. The increase in FLOPS is attributed to the inclusion of shallow FC layers (such as the $K$ and $Q$ layers) and a GRU module within the slot attention module. When this additional computational expense is juxtaposed against the variations in computational costs and parameter numbers of different backbones, we would say the extra overhead introduced by \ESCOUTER is almost negligible. This observation underscores the efficiency of \ESCOUTER, and the slight increase in FLOPS can be seen as a small trade-off for the significant benefits of model interpretability.

\subsection{Explainability}
\subsubsection{Quantitative Analysis} \label{quanti}

This section undertakes a detailed analysis to understand how the explainability of \ESCOUTER stands in comparison to other XAI methods. Table \ref{acc_tab} gives the scores of the metrics in Section \ref{sec:metrics} over CUB200 and ImageNet. Higher values ($\uparrow$) signify enhanced Precision and Insertion metrics. Conversely, lower values ($\downarrow$) indicate better performance for Deletion, Infidelity, Stability, and Time. To ensure robust analysis, we used 10,000 random samples in the ImageNet validation set and all validation samples in CUB200. We also evaluated SOTA positive explanation methods. These methods do not necessarily offer a negative variant. Hence, the comparison for the negative variant is over a small subset of the methods, including GradCAM \cite{selvaraju2017grad} and Saliency \cite{simonyan2013deep} that use the negative direction of gradients\footnote{In GradCAM, the negative direction of gradients is used to interpret the counterfactual samples (shown in Fig.~\ref{fig:counterfact}). It aims at explaining why an image does not belong to category $l$, which is consistent with our negative explanation.}.

\ESCOUTER demonstrates remarkable strengths for positive explanations. On ImageNet, \ESCOUTER achieves the highest Precision (0.7459 for $\lambda$=10), showing its ability to localize foreground objects. The Deletion score (0.1184 for $\lambda$=10) is also notably low. However, in terms of Insertion, \ESCOUTER is worse than GroupCAM, though still among the top. It also achieves the lowest Infidelity (0.3982 for $\lambda$=1), which shows \ESCOUTER's ability to explain the model decision even upon input perturbation faithfully. Meanwhile, \ESCOUTER gave lower Stability scores with a certain margin than gradient-based methods, showing that \ESCOUTER is more sensitive to small changes in input images. This is possibly caused by the overall attention of the slot computation, which counts noise more easily in the final decision-making. Similarly, on CUB200, \ESCOUTER exhibits the best Precision (0.7029 for $\lambda$=10), lowest Deletion (0.0495 for $\lambda$=3), and lowest Fidelity (0.4682 for $\lambda$=1), underscoring its effectiveness in discerning fine-grained features for bird species classification. From the Time metric, \ESCOUTER is more computationally expensive than the gradient-based methods because of the iterative update of the slot. Yet, it is faster than the perturbation-based methods.

\begin{figure}[t]
    \centering
    \begin{subfigure}[b]{1\columnwidth}
        \centering
        \includegraphics[width=0.8\columnwidth]{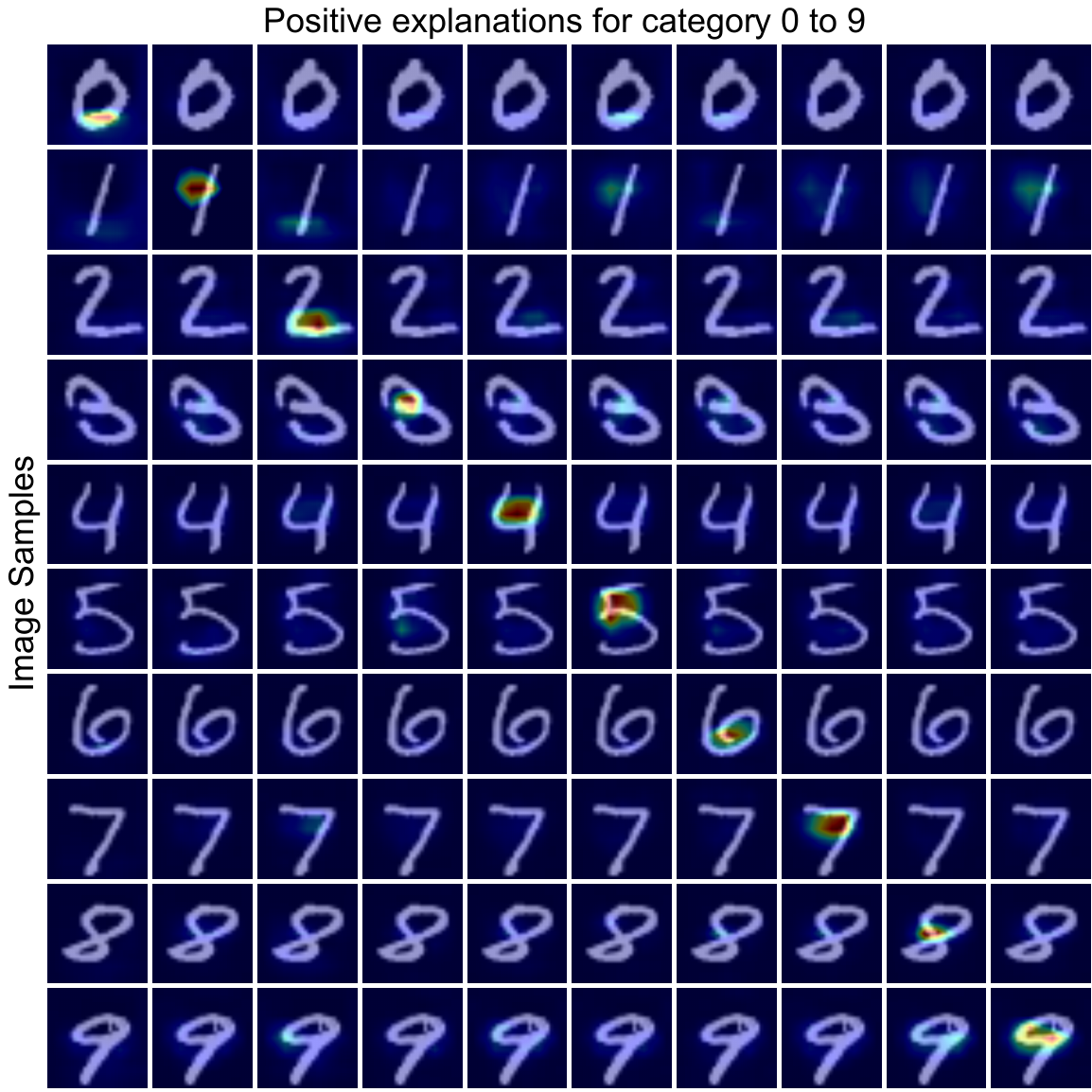}
        \caption{Explanation Confusion Matrix: why model predicts the images of [\textit{GT Category}] are [\textit{Predicted Category}].}
        \label{mnist_positive_matrix}
    \end{subfigure}
    \vspace{+0.1in}
    \begin{subfigure}[b]{1\columnwidth}
        \centering
        \includegraphics[width=0.8\columnwidth]{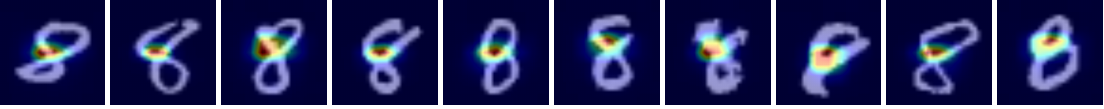}
        \caption{Explanation Consistency: why model predicts the images of a same category (``8") are ``8".}
        \label{mnist_positive_con}
    \end{subfigure}
    \caption{Positive supports for MNIST \cite{MNIST}. Using ResNet-18 \cite{he2016deep} as backbone and $\lambda$=10.}
    \label{mnist_postive}
\end{figure}

\begin{figure}[t]
    \centering
    \begin{subfigure}[b]{1\columnwidth}
        \centering
        \includegraphics[width=0.8\columnwidth]{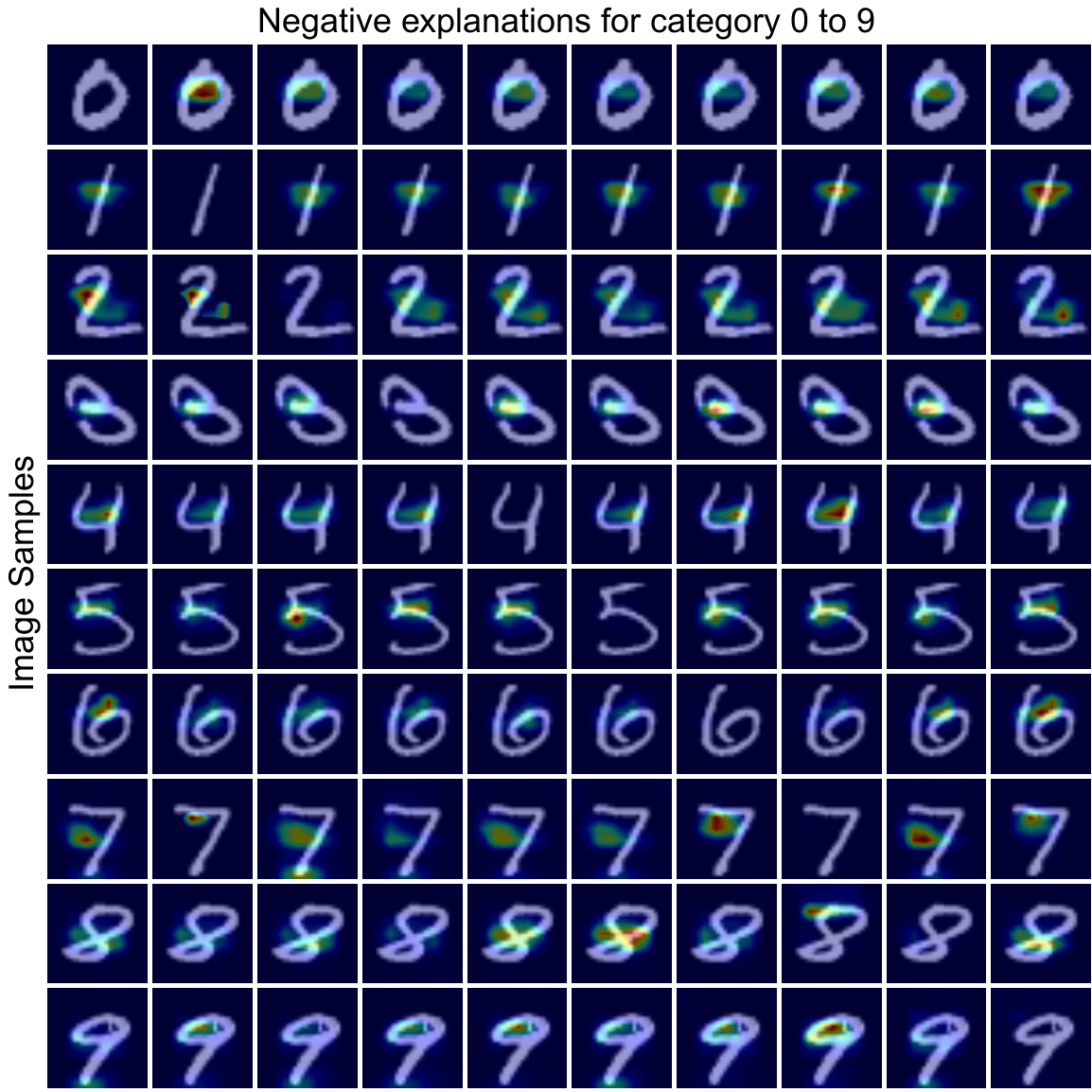}
        \caption{Explanation Confusion Matrix: why model not predicts the images of [\textit{GT Category}] are [\textit{Predicted Category}].}
        \label{mnist_negative_matrix}
    \end{subfigure}
    \vspace{+0.1in}
    \begin{subfigure}[b]{1\columnwidth}
        \centering
        \includegraphics[width=0.8\columnwidth]{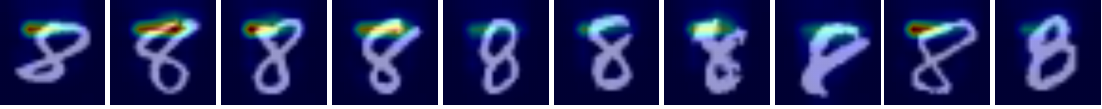}
        \caption{Explanation Consistency: why model predicts the images of the same category (``8'') are not ``7''.}
        \label{mnist_negative_con}
    \end{subfigure}
    \caption{Negative supports for MNIST \cite{MNIST}. Using ResNet-18 \cite{he2016deep} as backbone and $\lambda$=10.}
    \label{mnist_negative}
\end{figure}

We also observe that $\lambda$ influences the scores. Specifically, a larger value of $\lambda$ tends to reduce the size of the area $A$ as designed, which makes regions with higher attention weights more likely to fall into the object region $\bar{x}$, resulting in a higher Precision. 

There seems to be a trade-off between Insertion and Deletion. With a larger $\lambda$, \ESCOUTER tries to find supports within smaller regions. Consequently, deleting this area can effectively lower confidence. This behavior aligns with our design objective to identify supporting evidence for decisions. On the other hand, when computing Insertion, \ESCOUTER needs more pixels to regain confidence. This requirement stems from the slot's attention mechanism, which necessitates a view of enough pixels to retain the confidence. Stability is also decreased with an increase of $\lambda$. This implies that concentrating on a smaller region can reduce the impact of minor perturbations in breaking consistency between the model's output and explanation. However, limiting the size of the support region can result in a sluggish response to significant perturbations, often resulting in worse Infidelity scores.

\begin{figure*}[t]
\centering
\includegraphics[width=0.95\textwidth]{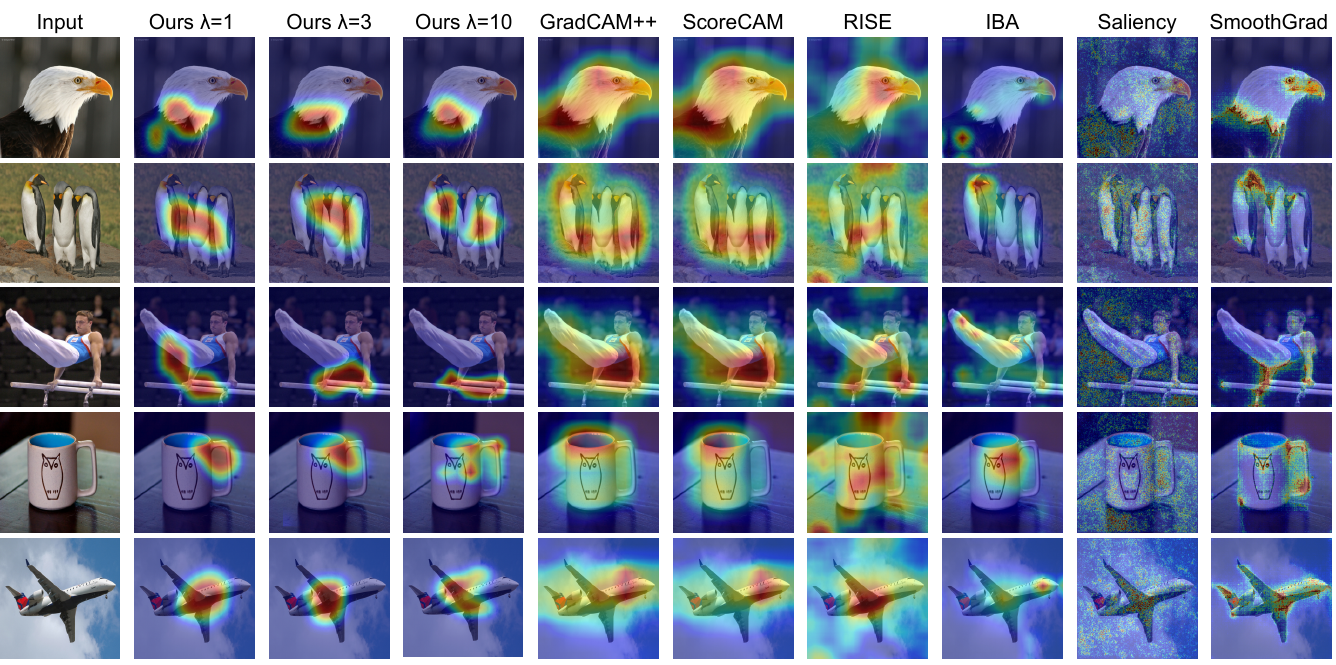}
\caption{Visualized positive explanations using \ESCOUTERpositive and existing XAI methods. We applied $\lambda$ equals to 1, 3, and 10 during the training process.}
\label{fig:positive_compare}
\end{figure*}

For negative explanations, \ESCOUTER achieves the highest Precision scores (0.7326 at $\lambda$=10). Notably, it records the lowest Deletion (0.1306 at $\lambda$=3) and Infidelity (0.3468 at $\lambda$=3). Compared to its positive variants, \ESCOUTERnegative demonstrates more significant improvements in performance. This is particularly evident in Precision, where it surpasses GradCAM++ by approximately 0.08. However, its weaknesses are still in Insertion and Stability. One notable distinction from \ESCOUTERpositive is the influence of $\lambda$. A larger $\lambda$ generally aids \ESCOUTER in deriving more effective explanations, but the best performance is achieved at $\lambda=3$. This means that excessively prioritizing the area loss can lead to sub-optimal explanations. These results indicate \ESCOUTER's sensitivity to $\lambda$.

In summary, across both datasets and variants, \ESCOUTER consistently outperforms the SOTA methods in Precision, Deletion, and Infidelity, while maintaining competitive scores in Insertion and Stability. This underscores its efficacy in providing interpretable and reliable explanations for model decisions. The time efficiency of \ESCOUTER is also notable, with it being relatively low compared to perturbation-based methods.

\subsubsection{Specificity and Consistency of Supports}
To better understand \ESCOUTER's interpretability, we visualized its attention weights $\bar{a}_l$, which serves as an explanation for category $l$. Our focus is three-fold: (i) identifying supports that \ESCOUTER relies on for decision-making, (ii) discerning how these supports vary across different categories, and (iii) examining their consistency within the same category. MNIST \cite{MNIST} is particularly suitable for this analysis due to the clear similarities and differences among its categories (i.e., digits), offering a simpler context than ImageNet and other natural image datasets. 

Firstly, we visualize $\bar{a}_l$ based on \ESCOUTERpositive for $l = \{\texttt{0}, \dots, \texttt{9}\}$ from left to right for a random input image of each category from top to down. We can see that the supports that \ESCOUTER found in the input image are specific to the category. Secondly, we give visualizations of $\bar{a}_l$ of some random input images of category $l$ (i.e., when $l$ is the ground-truth (GT) category $y$). They can provide some insights into the consistency of the supports. For this consistency analysis, we specifically examine the digit \texttt{8} with ten randomly selected samples from MNIST's test set.

From the visualizations in Fig.~\ref{mnist_positive_matrix}, it is evident that \ESCOUTERpositive adeptly identifies supports for images belonging to their ground-truth categories. Notably, strong attention is primarily observed only in the explanations for the ground-truth category, which is placed diagonally. This highlights \ESCOUTERpositive's effectiveness in focusing on category-specific features. Turning to the consistency analysis,  Fig.~\ref{mnist_positive_con} identifies the middle crossing part in digit \texttt{8} for all samples, which is a distinctive feature of this digit. This observation underscores \ESCOUTERpositive's ability to find consistent supports for the category.

Similarly, Fig.~\ref{mnist_negative_matrix} gives visualization for \ESCOUTERnegative. Unlike the positive variants, it does not identify supports for the ground-truth categories.  Instead, \ESCOUTERnegative consistently gives strong attention for non-ground-truth categories. Given the simplicity of digit recognition, \ESCOUTERnegative employs simple yet effective supports to refute most non-ground-truth categories. For instance, to distinguish an image of \texttt{1} from others, it consistently focuses on the central part of the vertical line because it can be a positive support for digit \texttt{1} and a negative support for the others.\footnote{The vertical line is also a negative support for digit \texttt{7}, which also has a slanted line. This support may respond to a line at a certain angle (i.e., closer to vertical) or may also cover the (absence of) a horizontal line of digit \texttt{7}.} When differentiating an image of \texttt{8} from digit \texttt{7}, the horizontal line is highlighted as the negative support. As shown in Fig.~\ref{mnist_negative_con}, this pattern of support is also consistently observed across other image samples of different digit \texttt{8}.

\subsubsection{Qualitative Analysis for Natural Images}
This section compares over natural images for \ESCOUTER and competing methods, including saliency-based methods (Saliency \cite{simonyan2013deep} and SmoothGrad \cite{smilkov2017smoothgrad}), gradient-based methods (GradCAM++ \cite{chattopadhay2018grad} and ScoreCAM \cite{wang2020score}), and perturbation-based methods (RISE \cite{petsiuk2018rise} and IBA \cite{IBA}).

\begin{figure*}[t]
    \centering
    \begin{subfigure}[b]{0.49\textwidth}
        \includegraphics[width=1\textwidth]{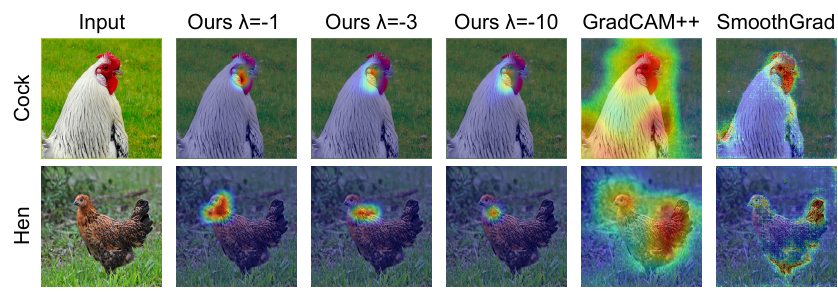}
        \caption{Negative support for \texttt{Cock} and \texttt{Hen}.}
        \label{negative_compare_1}
    \end{subfigure}
    \begin{subfigure}[b]{0.49\textwidth}
        \includegraphics[width=1\textwidth]{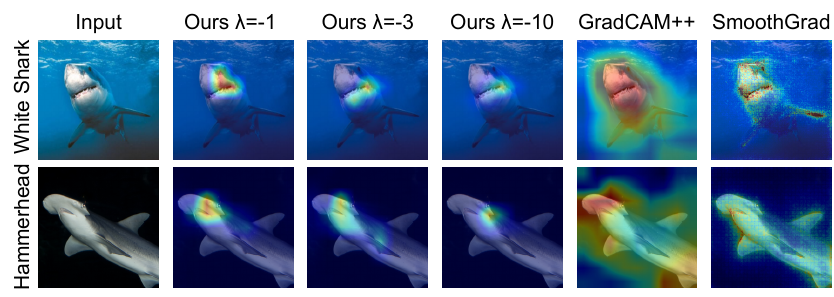}
        \caption{Negative support for \texttt{White Shark} and \texttt{Hammerhead}.}
        \label{negative_compare_2}
    \end{subfigure}
    \begin{subfigure}[b]{0.49\textwidth}
        \includegraphics[width=1\textwidth]{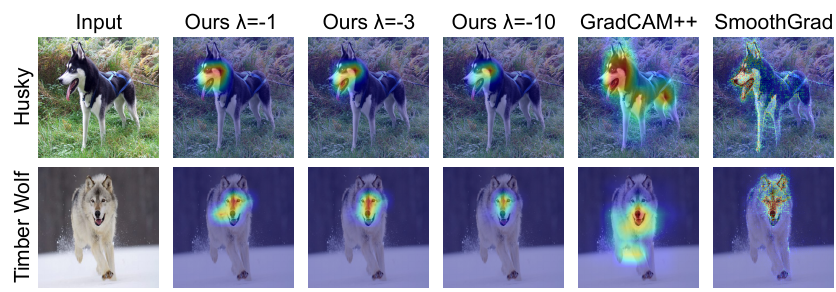}
        \caption{Negative support for \texttt{Husky} and \texttt{Timer Wolf}.}
        \label{negative_compare_3}
    \end{subfigure}
    \begin{subfigure}[b]{0.49\textwidth}
        \includegraphics[width=1\textwidth]{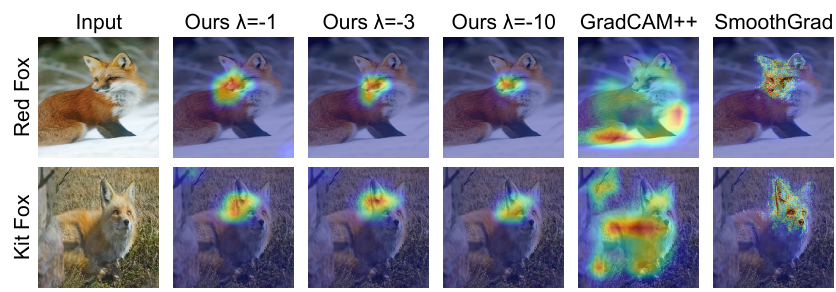}
        \caption{Negative support for \texttt{Red Fox} and \texttt{Kit Fox}.}
        \label{negative_compare_4}
    \end{subfigure}
    \caption{Visualized negative explanations using \ESCOUTERnegative and existing methods. We applied $\lambda$ equals to -1, -3, and -10 during the training process. Every image is marked on the left with its GT category label. The accompanying explanations are specifically designed to elucidate why an image of a certain GT category should not be classified into another category in the pair.}
    \label{negative_compare}
\end{figure*}

Fig.~\ref{fig:positive_compare} shows some examples of positive explanation by our method. They clearly illustrate that \ESCOUTERpositive effectively identifies supports that precisely cover the crucial features distinctive to each category. For instance, the \texttt{American Eagle} image, \ESCOUTERpositive highlights the black and white region at the neck. GradCAM++ and ScoreCAM also cover this area, encompassing a larger region. This trend is shared among other examples. For the \texttt{Horizontal Bar} image, our method mostly focuses on the part where the hand grips the pole. In the \texttt{Cup} image, it highlights the junction between the cup body and handle. Similarly, it identifies the part around the wing root for the \texttt{Air Plane} image. For an image with multiple target objects, as in the images in the second row from the top, \ESCOUTERpositive covers all penguins or even differentiates among them when $\lambda$ is higher (e.g., $\lambda$=10). It is also noteworthy that increasing $\lambda$ tends to reduce the size of the explanation area slightly. This constraint compels the model to seek out more fine-grained visual supports that are specific to each category.

\ESCOUTER also provides negative explanations, as demonstrated in Fig.~\ref{negative_compare}. We show pairs of images of closely related categories to highlight its capability to pinpoint negative supports. The ground-truth category for each image is shown on the left, while the negative supports for the other category are highlighted (e.g., Fig.~\ref{negative_compare_1}'s top image is of \texttt{Cock}, and the negative supports for \texttt{Hen} are highlighted in the images).  In Fig.~\ref{negative_compare_1}, \ESCOUTERnegative's negative support for the image of \texttt{Cock} not being \texttt{Hen} consistently focuses on the cock's face, particularly noting the red face as a key distinguishing characteristic. Conversely, when distinguishing the hen image from \texttt{Cock}, the model's attention is drawn to the neck feathers, likely influenced by the typically more lush feathers found around a cock's neck. This nuanced approach of \ESCOUTERnegative starkly contrasts with other methods like GradCAM, which tend to cover the entire foreground target, thus failing to isolate specific differentiating features. SmoothGrad, while has similar explanation region as \ESCOUTERnegative in its negative support for \texttt{Cock}, diverges in the case of negative support for \texttt{Hen} by focusing on less distinctive areas like the feet and tail.

\begin{figure}[t]
\centering
\includegraphics[width=1\columnwidth]{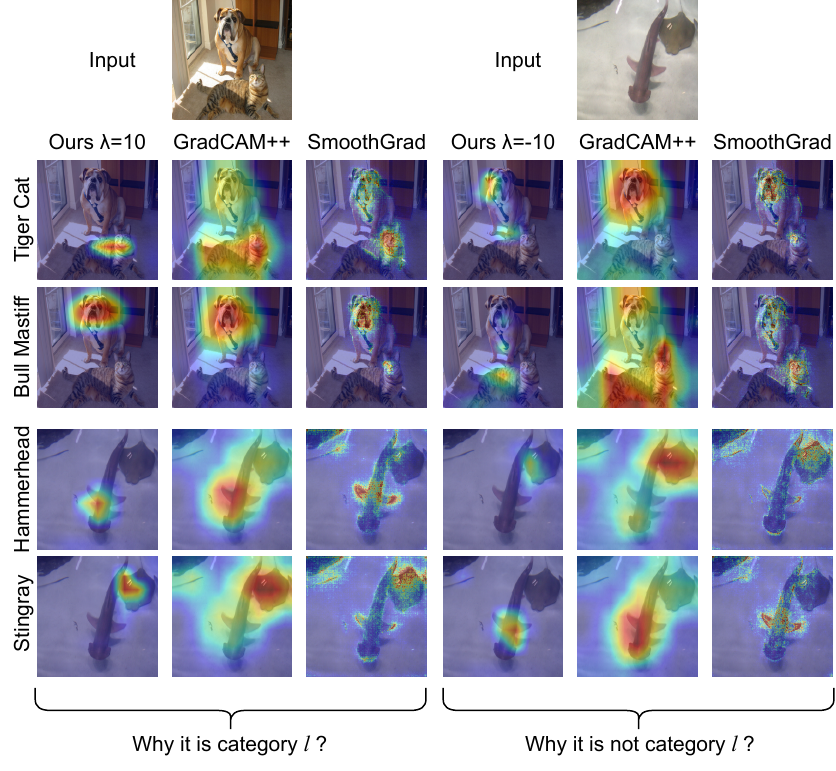}
\caption{Visualization of positive and negative supports for counterfactual samples. Each image is marked on the left with its assumed target category label.}
\label{fig:counterfact}
\vspace{-0.15in}
\end{figure}

\begin{table*}[t!]
    \caption{Performance comparison of \ESCOUTER and its variants on ImageNet dataset. \(\lambda\) is set to \(10\) during training and ResNet-50 is adopted as the backbone. The explanation performance is measured on the GT category for the positive explanation and on the least similar class (LSC) for the negative explanation.}
    \label{table_exp_ablation_study}
    \centering
    \resizebox{.89\textwidth}{!}{%
        \begin{tabular}{c|c|cc|c|ccccc}
            \toprule
            \multirow{2}{*}{Exp. Type} & \multirow{2}{*}{Variants} & \multicolumn{2}{c|}{Computational Costs} & Classification & \multicolumn{5}{c}{Explainability} \\
            & & Params (M) & Flops (G) & Accuracy & Pre. $\uparrow$ & Ins. $\uparrow$ & Del. $\downarrow$ & Inf. $\downarrow$ & Sta. $\downarrow$ \\
            \midrule
            \multirow{3}{*}{Positive} & \ESCOUTER$^+$ & 23.7406 & 6.1016 & \underline{\textbf{0.7595}} & \underline{\textbf{0.7459}} & \underline{\textbf{0.6802}} & \underline{\textbf{0.1184}} & \underline{\textbf{0.4412}} & \underline{\textbf{0.0987}} \\
            & \textit{w.o. GRU} & \underline{\textbf{23.3527}} & \underline{\textbf{6.0755}} & 0.7548 & 0.7319 & 0.6743 & 0.1232 & 0.4650 & 0.1068 \\
            & \textit{w.o. PE} & 23.7406 & 6.1016 & 0.7514 & 0.7283 & 0.6704 & 0.1292 & 0.4742 & 0.1132 \\
            \midrule
            \multirow{3}{*}{Negative} & \ESCOUTER$^-$ & 23.7406 & 6.1016 & \underline{\textbf{0.7507}} & \underline{\textbf{0.7326}} & \underline{\textbf{0.6107}} & \underline{\textbf{0.1450}} & \underline{\textbf{0.4514}} & \underline{\textbf{0.0752}} \\
            & \textit{w.o. GRU} & \underline{\textbf{23.3527}} & \underline{\textbf{6.0755}} & 0.7460 & 0.7304 & 0.6059 & 0.1515 & 0.4827 & 0.1003 \\
            & \textit{w.o. PE} & 23.7406 & 6.1016 & 0.7413 & 0.7317 & 0.6001 & 0.1567 & 0.4902 & 0.0936 \\ 
            \bottomrule
        \end{tabular}
    }
\end{table*}

\begin{table}[!t]
	\caption{Area sizes of the explanations based on category similarity. Calculation is implemented over 5,000 samples randomly selected from the validation set of ImageNet.}
	\label{table_exp_pos_neg_compare}
	\centering
	\resizebox{1\columnwidth}{!}{%
    	\begin{tabular}{lcccc}
    		\toprule
    	    \multirow{2}{*}{Methods} & \multicolumn{4}{c}{Target Categories}\\
            \cmidrule(lr){2-5}
    	    &GT  &  Highly-similar & Similar  & Dissimilar \\
    	    \midrule
            \ESCOUTER$^+$$_{\lambda=10}$ & .2863 & .1291 & .0714 & .0373\\
            \ESCOUTER$^-$$_{\lambda=10}$ & .0116 & .0225 & .0572 & .1704\\
    		\bottomrule
    	\end{tabular}
	}
\end{table}

Fig.~\ref{negative_compare_2} also illustrates how \ESCOUTERnegative distinguishes different breeds of sharks. The teeth serve as the primary feature for differentiating \texttt{White Shark} from \texttt{Hammer Headed}, where the converse situation is identified mainly by the head part. Similarly, Fig.~\ref{negative_compare_3} demonstrates that \ESCOUTERnegative consistently focuses on the face to distinguish between \texttt{Husky} and \texttt{Timber Wolf}. Moreover, in Fig.~\ref{negative_compare_4}, the distinction is made evident as highlighting the face for \texttt{Red Fox} and the ears for \texttt{Kit Fox}. These examples showcase the precision of \ESCOUTERnegative in identifying specific features crucial for differentiating similar categories. In addition, it is evident that increasing the value of $\lambda$ effectively narrows the focus of the attention area, thereby improving the ability to pinpoint negative supports (refer to Table \ref{acc_tab}).

To further showcase \ESCOUTER's effectiveness in simultaneously identifying both positive and negative supports in a single image, we present examples involving two objects. The first image (the left image at the top) contains both \texttt{Tiger Cat} (a cat) and \texttt{Bull Mastiff} (a dog). The three images from the left in the row are positive explanations for the category on the left, and the other three images are negative explanations for the same category. \ESCOUTER excels in locating the foreground object in its positive and negative explanation. For instance, in the positive explanation, \ESCOUTERpositive accurately identifies the heads of both the dog and cat when explaining ``why it is category $l$''. Similarly, in the negative explanation, it also highlights the head region to illustrate ``why it is not category $l$". GradCAM++ and SmoothGrad provide both positive and negative explanations, but they are not designed to pinpoint discerning features. For example, GradCAM++ seems to spot the foreground objects, encompassing both the cat and dog for \texttt{Tiger Cat}. This tendency is also seen in the second example, showing both \texttt{Hammer Headed} and \texttt{Stingray}, where \ESCOUTER successfully distinguishes them. These results demonstrate \ESCOUTER's ability to learn fine-grained explanations in complex scenarios where multiple categories appear simultaneously.

\subsubsection{Semantic Similarity of Categories and Explanations}

\ESCOUTER is designed to give a larger relevance score (as an explanation) to distinguishing features of the ground-truth category (or of non-ground-truth in the negative variant), while showing a smaller value to other regions. It is interesting to explore how \ESCOUTER reacts to semantically similar categories. We make pairs of categories based on their similarity scores defined in Eq.~(\ref{eq_wordnet_similarity}). Specifically, if the similarity between a pair of categories is equal to or higher than $0.9$, the pair is marked as `highly similar.' If it is equal to or higher than 0.7 but less than 0.9, it is marked as `similar,' and any other pairs are considered `dissimilar.' To quantify the \textit{strength of explanation} (SE), we use the sum of relevance scores, formulated as 
\begin{equation}
    \text{SE}_l(x) = \sum_{p \in x} r_p^l(x) / |x|,
\end{equation}
where we extend the notation of relevance score $r_p(x)$ to $r_p^l(x)$ to clarify the category, and $|x|$ gives the number of pixels in $x$. 

Table~\ref{table_exp_pos_neg_compare} summarizes SE evaluated over the image of a category for the other category of highly similar, similar, and dissimilar category pairs. Note that SE is averaged over 5,000 samples randomly selected from the validation set of ImageNet. A distinct trend emerges from these results: SE for \ESCOUTERpositive decreases as the semantic similarity decreases. Conversely, \ESCOUTERnegative tends to yield larger SE for categories that are less similar. This adaptive power to explanatory regions based on category similarity is meaningful to application scenarios requiring globally interpretable measures. 

\subsection{Ablation Study}\label{ablation}

This section quantifies the impact of GRU and PE, while retaining the others unchanged.

To this end, we evaluate two variants: One is \ESCOUTER without GRU (i.e., $T = 0$). The other variant is without PE, feeding flattened input features to \ESCOUTER without positional information. These variants are trained and evaluated over ImageNet with ResNet-50 backbone and $\lambda = 10$.  Table~\ref{table_exp_ablation_study} shows the metrics. We observed that incorporating GRU incurs a minor increase in the number of parameters and FLOPs. PE does not change computational metrics. Notably, the exclusion of GRU leads to a decline in classification accuracy. Removal of PE also worsens the accuracy. Moreover, the table clearly demonstrates that GRU and PE contribute to the XAI metrics. We can conclude that the absence of GRU and PE not only diminishes classification performance but also adversely affects all aspects of explainability, underscoring their vital roles in \ESCOUTER.

\begin{figure}[t]
\centering
\includegraphics[width=1\columnwidth]{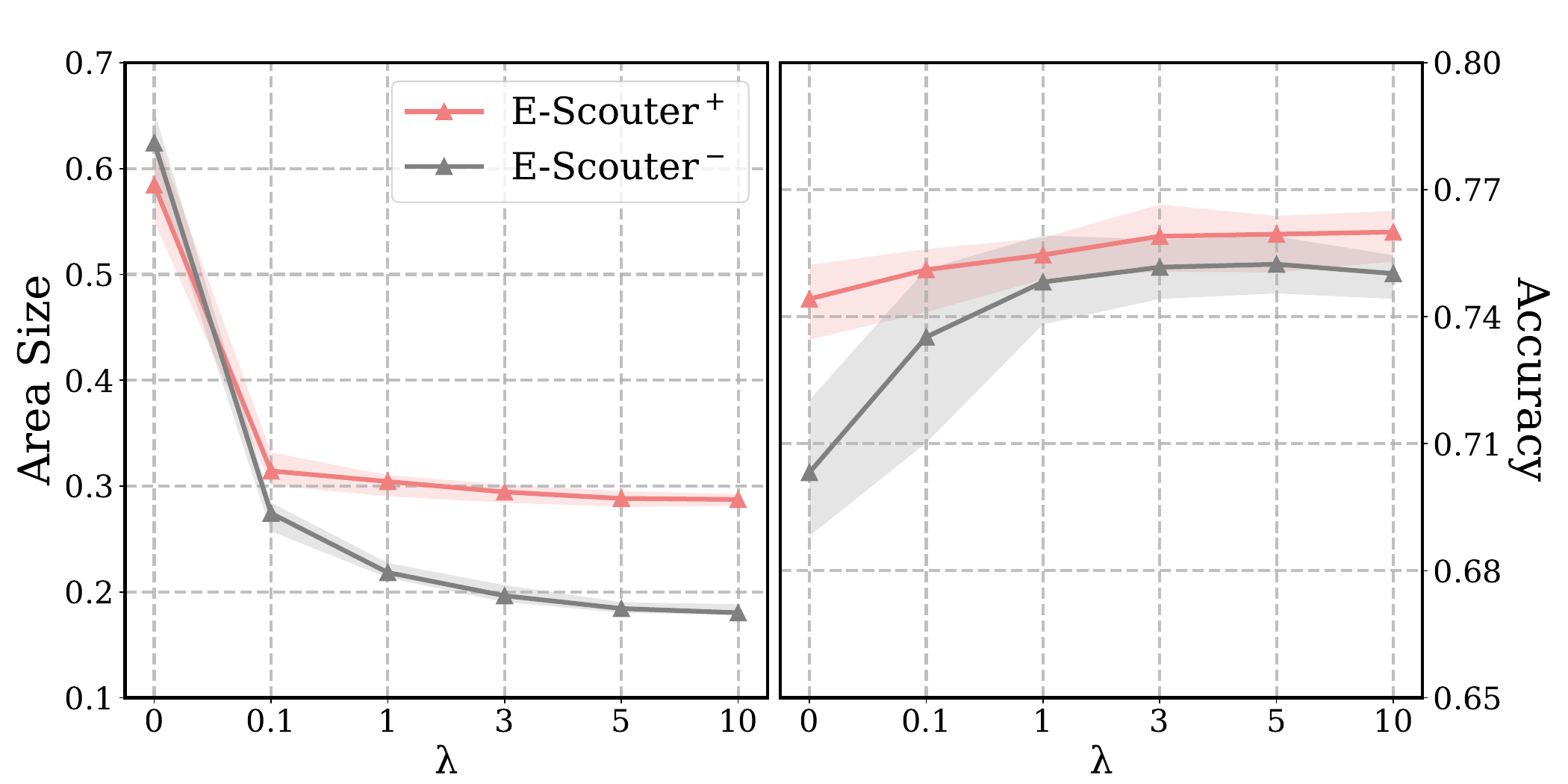}
\caption{Classification and area size variance with different value of $\lambda$. Experiments are implemented on ImagegNet and use ResNet-50 as backbone.}
\label{fig:acc_size}
\end{figure}

\begin{table}[!t]
    \caption{Comparison between \ESCOUTER and SCOUTER in different XAI metrics. All experiments are implemented on ImageNet (200 categories) with ResNet-50 as a backbone and $\lambda$=10. }
    \label{tab:compare_previosu}
    \centering
    \resizebox{0.89\columnwidth}{!}{%
        \begin{tabular}{lccccc}
            \toprule
            \multirow{2}{*}{Method} & \multicolumn{5}{c}{Explainability} \\
             & Pre. $\uparrow$ & Ins. $\uparrow$ & Del. $\downarrow$ & Inf. $\downarrow$ & Sta. $\downarrow$ \\
            \midrule
            \multirow{1}{*}{\ESCOUTER$^+$}  & .7392 & \underline{\textbf{.7240}} & .1218 & \underline{\textbf{.4286}} & \underline{\textbf{.1045}} \\
            \multirow{1}{*}{SCOUTER$^+$}  & \underline{\textbf{.7520}} & .7085 & \underline{\textbf{.1132}} & .4325 & .1426 \\
            \cmidrule(lr){1-6}
            \multirow{1}{*}{\ESCOUTER$^-$}  & .7388 & \underline{\textbf{.6573}} & .1309 & \underline{\textbf{.4032}} & \underline{\textbf{.0765}} \\
            \multirow{1}{*}{SCOUTER$^-$}  & \underline{\textbf{.7404}} & .6456 & \underline{\textbf{.1286}} & .4217 & .0947 \\
            \bottomrule
        \end{tabular}
    }
\end{table}

We also investigated the influence of the parameter $\lambda$ on the size of explanatory regions and classification accuracy. We train the model five times for each value of $\lambda$. Fig.~\ref{fig:acc_size} (left) demonstrates the influence of $\lambda$ on the size of explanatory regions, in line with our area loss design. The size is quantified similarly to the area loss in Eq.~(\ref{eq:area_loss}) but only for the ground-truth categories (LSC for \ESCOUTERnegative). Formally, using the notation $r_p(x)$, the area size is defined as:
\begin{equation}
    \text{Area size} = \sum_{p \in x} r_p(x) / |x|,
\end{equation}
where $|x|$ gives the number of pixels in $x$. Setting $\lambda$ to 0 results in a larger area size. As $\lambda$ increases, the area size decreases.  

Fig.~\ref{fig:acc_size} (right) illustrates the variation in accuracy. Our findings are two-fold: (i) Higher values of $\lambda$ tend to improve accuracy for both the positive and negative variants. (ii) A smaller attention region aids the model in identifying significant features, particularly in \ESCOUTERnegative. {Reasons here.} It is also noteworthy that the accuracy of \ESCOUTERnegative is initially more diverse, and it becomes more stable as $\lambda$ increases. It is important to note the possible trade-off involved: as depicted in Figure \ref{fig:acc_size} (right), larger $\lambda$ may adversely affect classification performance as we can see a slight decrease in accuracy for \ESCOUTERnegative with $\lambda$=10. Furthermore, as discussed in Section \ref{quanti}, explainability also declines with higher $\lambda$. In conclusion, $\lambda$ emerges as a crucial hyper-parameter for \ESCOUTER. Tuning $\lambda$ for a target dataset can be the key to achieving a balance between higher accuracy and better explainability.

\begin{figure}[t]
\centering
\includegraphics[width=1\columnwidth]{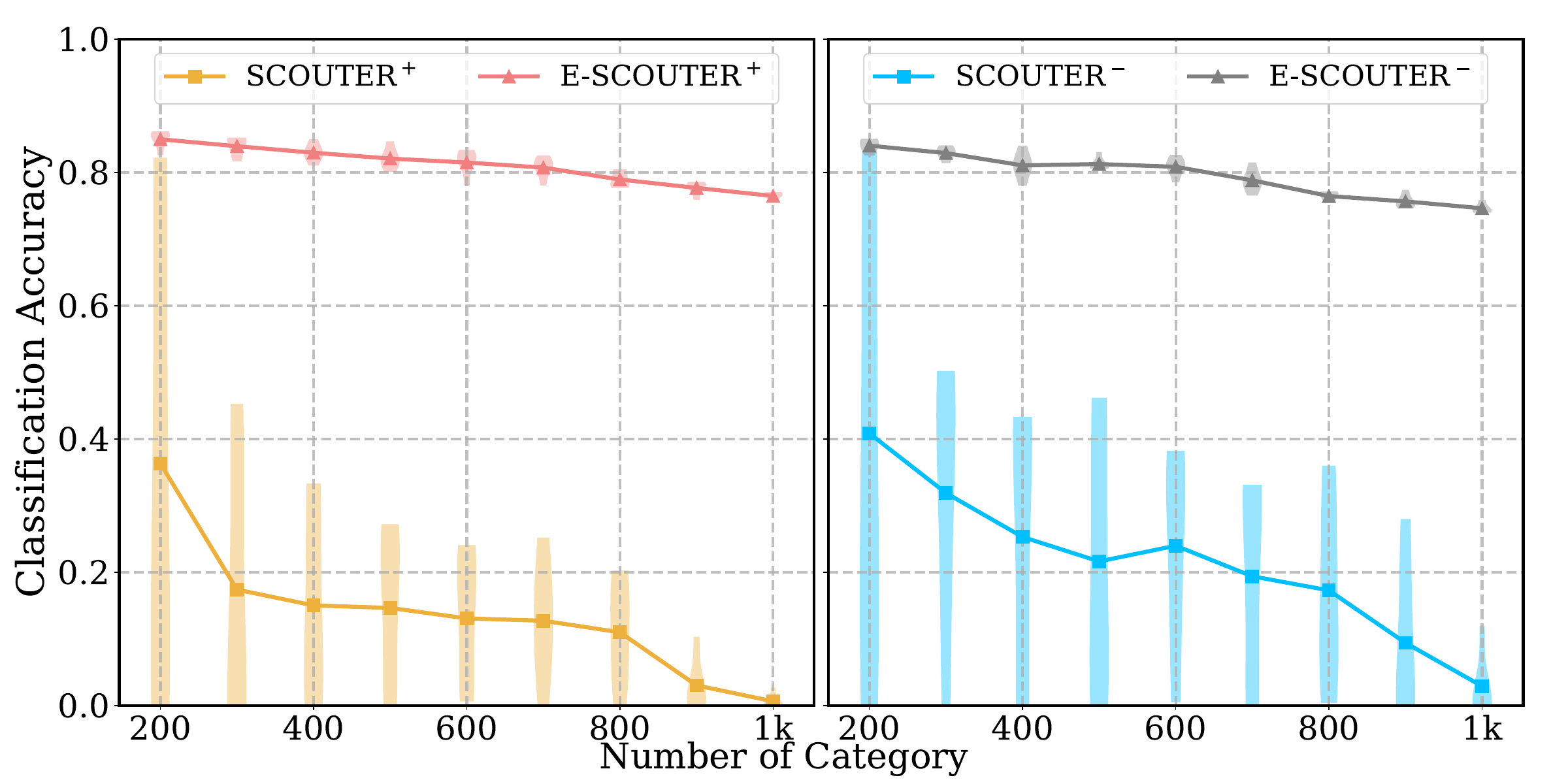}
\caption{Comparison of classification accuracy between \ESCOUTER and SCOUTER across different category number for training ImageNet. We adopt 5 times training for each category number setting (using ResNet-50 and $\lambda$=10).}
\label{fig:to_previous}
\end{figure}

\subsection{Comparison to SCOUTER} \label{compare_previous}
We compare \ESCOUTER to the original SCOUTER. In this updated version, we have made two key improvements.

\textbf{Stability in training with more categories :} Incorporating a normalization step (refer to Eq.~\ref{eq:normalization}) has significantly stabilized training, enabling \ESCOUTER to handle more than 200 categories, allowing for its utility in various scenarios. To experimentally show this, we trained the model five times for each category number, presenting violin plots with line plots of the mean values. The improvement is evident in Fig.~\ref{fig:to_previous}, where the classification accuracies of \ESCOUTERpositive (red line) and SCOUTER$^+$ (yellow line) are depicted on the left and of \ESCOUTERnegative (grey line) and SCOUTER$^-$ (blue line) on the right. The plots clearly show a gradual decrease in \ESCOUTER's accuracy. As discussed in Section \ref{classifier}, \ESCOUTER's performance is on par with the FC classifier. SCOUTER achieves high accuracy for 200 categories in some training attempts, but its performance significantly deteriorates with more categories. Notably, SCOUTER experiences training failures (the accuracy is stuck to almost the chance rate) even with fewer categories, while \ESCOUTER maintains stability up to 1,000 categories. We found that our new normalization can prevent slots from generating excessive attention values, which may approximate hard attention and lead to training failures,  stabilizing the training. 

\textbf{Higher Interpretability:} \ESCOUTER not only excels in terms of accuracy but also exhibits superior interpretability in both negative and positive versions. This enhanced interpretability is crucial, as it provides deeper insights into the decision-making process of the model, making it more transparent and trustworthy. As illustrated in Table \ref{tab:compare_previosu} (only using 200 categories for training), when evaluated against the five key XAI metrics, \ESCOUTER outperforms SCOUTER (slightly lower in Precision and Delation), maintaining a noticeable margin. These findings underscore the substantial improvements from \ESCOUTER.

\subsection{Case Study on Medical Image Diagnosis} \label{case}

\begin{figure}[t]
\centering
\includegraphics[width=0.8\columnwidth]{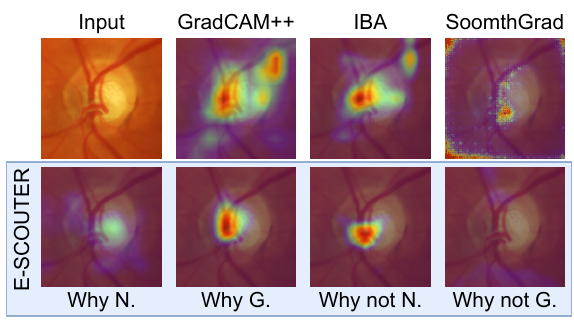}
\caption{Explanations for a positive sample in the glaucoma diagnosis dataset. Top row shows the input image and previous XAI methods explanations. Bottom row presents \ESCOUTERpositive (first and second columns) and \ESCOUTERnegative (third and fourth columns) explanations for Normal (N.) and Glaucoma (G.) cases, using $\lambda=10$.}
\label{fig:retina}
\end{figure}

\begin{figure}[t]
\centering
\includegraphics[width=0.98\columnwidth]{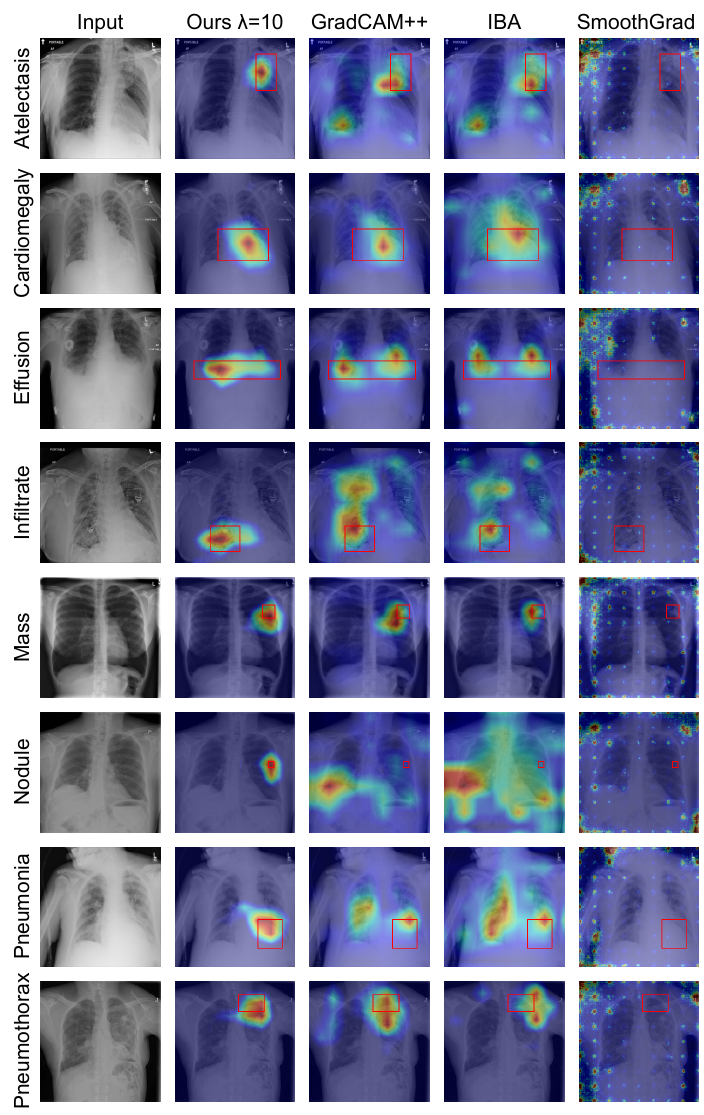}
\caption{Explanations for samples from the Xray-14 \cite{wang2017chestx} dataset. The results are compared among \ESCOUTER and other methods. The disease name for each input sample is listed on the left and the lesion area is annotated with a red bounding box.}
\label{fig:xray}
\end{figure}

\ESCOUTER incorporates an area loss feature that limits the size of the support region. This is particularly advantageous in applications like medical image classification, where smaller, more focused support regions can accurately highlight symptoms, providing greater informative value. Notably, existing methods lacked the capability to furnish negative explanations --- a critical need in medical diagnostics, where understanding the rationale for ruling out certain conditions is as important as diagnosing them. \ESCOUTER is able to address this for applications such as glaucoma detection (see Fig.~\ref{fig:retina}) and chest X-ray analysis (refer to Fig.~\ref{fig:xray}). In Table \ref{table_case_study}, we present a comparative analysis of both \ESCOUTER variants against the FC classifier, utilizing two datasets. The comparison metrics include Area Under the Curve (AUC), Accuracy (Acc.), F1-score, and an XAI metric Precision (Pre.). For the ACRIMA dataset \cite{ACRIMA}, the Precision metric is based on the foreground region of the Optic Cup (OC), as done by the REFUGE \cite{orlando2020refuge}. In the case of the X-ray dataset, Precision is determined using lesion annotations provided by the Xray-14 dataset \cite{wang2017chestx}.

\begin{table}[!t]
	\caption{Classification Performance on ACRIMA \cite{ACRIMA} and X-ray14 \cite{wang2017chestx} datasets. ResNet-50 is adopted as the backbone and $\lambda$ is defaulted as 10. For FC, we use GrandCAM++ \cite{chattopadhay2018grad} to compute the Precision (Pre.).}
	\label{table_case_study}
	\centering
	\resizebox{0.96\columnwidth}{!}{%
    	\begin{tabular}{clcccc}
    		\toprule
    	     Dataset & Classifier &  AUC & Acc. & F1 & Pre.\\
    	    \midrule
    	    \multirow{3}{*}{ACRIMA \cite{ACRIMA}} & FC & .9992 & .9921 & .9866 & .3407 \\
    	    & \ESCOUTER$^+$ & \underline{\textbf{1.000}} & \underline{\textbf{1.000}} & \underline{\textbf{1.000}} & .4852 \\
    	    & \ESCOUTER$^-$ & .9997 & .9929 & .9932 &  \underline{\textbf{.5008}}\\
            \cmidrule(lr){2-6}
            \multirow{3}{*}{Xray-14 \cite{wang2017chestx}} & FC & .7787 & .9453 & .8525 & .1858 \\
    	    & \ESCOUTER$^+$ & \underline{\textbf{.7801}} & \underline{\textbf{.9460}} & \underline{\textbf{.8560}} & \underline{\textbf{.2549}}\\
    	    & \ESCOUTER$^-$ & .7746 & .9405 & .8427 & .2316\\
    		\bottomrule
    	\end{tabular}
	}
\end{table}

We deployed \ESCOUTER with a $\lambda=10$ on the ACRIMA dataset \cite{ACRIMA}, which categorizes images into \texttt{Normal} and \texttt{Glaucoma} categories. Utilizing a ResNet-50 architecture as the backbone, we observed better classification performance from both \ESCOUTERpositive and \ESCOUTERnegative models compared to the FC classifier, as detailed in Table \ref{table_case_study}. Our method also has much better Precision to locate the OC area. Of particular note is the preference for \ESCOUTER in clinical settings, where doctors value precise identification of regions in the optic disc. As illustrated in Fig.~\ref{fig:retina}, \ESCOUTER’s visualizations pinpoint clinically meaningful features, such as changes in vessel shape associated with optic cup enlargement. While the IBA model also identifies small regions, it often includes irrelevant or non-informative areas. In this context, \ESCOUTER’s capability to provide fine-grained negative explanations proves particularly beneficial, especially when machine diagnoses diverge from doctors’ initial assessments.

\ESCOUTER exhibits outstanding performance on the Xray-14 dataset, with the positive variant surpassing the FC classifier across all classification metrics. Although there is a slight decrease in performance in the negative variant, it still remains comparable. A key strength of our method lies in its ability to align the explanation area closely with lesion annotations. As indicated in Table \ref{table_case_study}, both variants achieve higher Precision compared to GradCAM++. Fig.~\ref{fig:xray} showcases a variety of visualization samples for each disease category, where it is evident that \ESCOUTER typically generates smaller explanation areas that more accurately coincide with the lesion annotations (indicated by red bounding boxes). This is particularly notable in complex diagnoses, such as nodules, where other methods struggle to identify compact regions as decisive factors. This underscores \ESCOUTER's effectiveness in discriminative localization \cite{CAM}. In conclusion, our approach not only captures critical lesion features but also aligns more consistently with medical professionals' analyses.

\section{Discussion}
We present \ESCOUTER, a visual explainable classifier. This method distinguishes itself through its robust performance across diverse explainability metrics and datasets.

As a classifier, \ESCOUTER showcases competitive performance in comparison to traditional FC classifiers. Our extensive experiments demonstrate that \ESCOUTER, in both its positive and negative variants, consistently maintains high accuracy levels across a range of datasets and backbone models. These results affirm \ESCOUTER's effectiveness as a classifier. Moreover, its contribution lies in the explainability. In recent years, XAI has emerged as a critical focus, with the understanding of the rationale behind AI decisions becoming increasingly important, particularly in high-stakes domains. \ESCOUTER propels this field forward by offering explanations that are not only more precise but also more reflective of the model's actual decision-making process. Additionally, by employing our unique attention area loss, \ESCOUTER gains the capability to regulate the size of explanatory regions. This feature enables \ESCOUTER to adaptively align with specific task requirements, providing accurate explanations for model decisions.

\ESCOUTER's capability for negative explanation is another key innovation. It adeptly differentiates between similar categories by identifying negative supports in an image. This ability to elucidate the ``why not'' behind a classification decision is especially valuable in scenarios requiring fine-grained classification. Furthermore, negative explanations can be invaluable for machine teaching. Understanding the reasons behind a model's exclusion of certain categories can be as crucial as comprehending its final decision.

In the domain of medical image analysis, \ESCOUTER's capacity to provide explanations is exceptionally beneficial. As discussed in Section \ref{case}, it pinpoints lesion areas in medical images, a critical aspect for diagnostic accuracy. The precision with which \ESCOUTER identifies and elucidates specific regions within an image equips healthcare professionals with a deeper understanding of AI-based diagnostic decisions, thus bolstering trust in AI-assisted medical diagnosis. In summary, \ESCOUTER signifies a leap forward in XAI for image recognition. Its dual capability of providing both positive and negative explanations, alongside its competitive performance relative to traditional classifiers, and its potential applicability in crucial areas, like medical imaging, render it an important step forward in the XAI field.

\section{Conclusion}
In this paper, we introduced a visually explainable classifier, \ESCOUTER. Built upon slot attention, it enhances the model's ability to focus on relevant image regions and ensures that the explanations provided align with the model's internal reasoning processes. Our extensive experimental evaluations have demonstrated that our method not only excels in providing accurate and meaningful explanations of its classification decisions but also maintains a high level of classification performance. These results attest to the effectiveness of \ESCOUTER as a SOTA tool in the field of XAI. We believe it contributes to the ongoing discourse in the AI community regarding explainability and sets a new benchmark for future developments.

\bibliographystyle{IEEEtran}
\bibliography{mybib}

\begin{thebibliography}{10}
\providecommand{\url}[1]{#1}
\csname url@samestyle\endcsname
\providecommand{\newblock}{\relax}
\providecommand{\bibinfo}[2]{#2}
\providecommand{\BIBentrySTDinterwordspacing}{\spaceskip=0pt\relax}
\providecommand{\BIBentryALTinterwordstretchfactor}{4}
\providecommand{\BIBentryALTinterwordspacing}{\spaceskip=\fontdimen2\font plus
\BIBentryALTinterwordstretchfactor\fontdimen3\font minus \fontdimen4\font\relax}
\providecommand{\BIBforeignlanguage}[2]{{%
\expandafter\ifx\csname l@#1\endcsname\relax
\typeout{** WARNING: IEEEtran.bst: No hyphenation pattern has been}%
\typeout{** loaded for the language `#1'. Using the pattern for}%
\typeout{** the default language instead.}%
\else
\language=\csname l@#1\endcsname
\fi
#2}}
\providecommand{\BIBdecl}{\relax}
\BIBdecl

\bibitem{amann2020explainability}
J.~Amann, A.~Blasimme, E.~Vayena, D.~Frey, and V.~I. Madai, ``Explainability for artificial intelligence in healthcare: a multidisciplinary perspective,'' \emph{BMC medical informatics and decision making}, vol.~20, no.~1, pp. 1--9, 2020.

\bibitem{van2022explainable}
B.~H. Van~der Velden, H.~J. Kuijf, K.~G. Gilhuijs, and M.~A. Viergever, ``Explainable artificial intelligence (xai) in deep learning-based medical image analysis,'' \emph{Medical Image Analysis}, vol.~79, p. 102470, 2022.

\bibitem{reddy2022explainability}
S.~Reddy, ``Explainability and artificial intelligence in medicine,'' \emph{The Lancet Digital Health}, vol.~4, no.~4, pp. e214--e215, 2022.

\bibitem{simonyan2013deep}
K.~Simonyan, A.~Vedaldi, and A.~Zisserman, ``Deep inside convolutional networks: Visualising image classification models and saliency maps,'' in \emph{Workshop, ICLR}, 2013.

\bibitem{selvaraju2017grad}
R.~R. Selvaraju, M.~Cogswell, A.~Das, R.~Vedantam, D.~Parikh, and D.~Batra, ``Grad-{CAM}: Visual explanations from deep networks via gradient-based localization,'' in \emph{CVPR}, 2017, pp. 618--626.

\bibitem{petsiuk2018rise}
V.~Petsiuk, A.~Das, and K.~Saenko, ``{RISE}: {R}andomized input sampling for explanation of black-box models,'' in \emph{BMVC}, 2018.

\bibitem{wang2020scout}
P.~Wang and N.~Vasconcelos, ``Scout: Self-aware discriminant counterfactual explanations,'' in \emph{CVPR}, 2020, pp. 8981--8990.

\bibitem{li2021scouter}
L.~Li, B.~Wang, M.~Verma, Y.~Nakashima, R.~Kawasaki, and H.~Nagahara, ``Scouter: Slot attention-based classifier for explainable image recognition,'' in \emph{ICCV}, 2021, pp. 1046--1055.

\bibitem{MNIST}
Y.~LeCun, L.~Bottou, Y.~Bengio, and P.~Haffner, ``Gradient-based learning applied to document recognition,'' \emph{Proceedings of the IEEE}, vol.~86, no.~11, pp. 2278--2324, 1998.

\bibitem{ImageNet}
J.~{Deng}, W.~{Dong}, R.~{Socher}, L.-J. {Li}, {Kai Li}, and {Li Fei-Fei}, ``{ImageNet}: A large-scale hierarchical image database,'' in \emph{CVPR}, 2009, pp. 248--255.

\bibitem{CUB-200}
P.~Welinder, S.~Branson, T.~Mita, C.~Wah, F.~Schroff, S.~Belongie, and P.~Perona, ``{Caltech-UCSD Birds 200},'' California Institute of Technology, Tech. Rep. CNS-TR-2010-001, 2010.

\bibitem{chang2021explaining}
J.~Chang, J.~Lee, A.~Ha, Y.~S. Han, E.~Bak, S.~Choi, J.~M. Yun, U.~Kang, I.~H. Shin, J.~Y. Shin \emph{et~al.}, ``Explaining the rationale of deep learning glaucoma decisions with adversarial examples,'' \emph{Ophthalmology}, vol. 128, no.~1, pp. 78--88, 2021.

\bibitem{wang2020score}
H.~Wang, Z.~Wang, M.~Du, F.~Yang, Z.~Zhang, S.~Ding, P.~Mardziel, and X.~Hu, ``Score-cam: Score-weighted visual explanations for convolutional neural networks,'' in \emph{CVPRW}, 2020, pp. 24--25.

\bibitem{locatello2020object}
F.~Locatello, D.~Weissenborn, T.~Unterthiner, A.~Mahendran, G.~Heigold, J.~Uszkoreit, A.~Dosovitskiy, and T.~Kipf, ``Object-centric learning with slot attention,'' \emph{NeurIPS}, vol.~33, pp. 11\,525--11\,538, 2020.

\bibitem{CAM}
B.~Zhou, A.~Khosla, A.~Lapedriza, A.~Oliva, and A.~Torralba, ``Learning deep features for discriminative localization,'' in \emph{CVPR}, 2016, pp. 2921--2929.

\bibitem{chattopadhay2018grad}
A.~Chattopadhay, A.~Sarkar, P.~Howlader, and V.~N. Balasubramanian, ``Grad-cam++: Generalized gradient-based visual explanations for deep convolutional networks,'' in \emph{WACV}, 2018, pp. 839--847.

\bibitem{IBA}
S.~Karl, S.~Leon, T.~Federico, and L.~Tim, ``Restricting the flow: Information bottlenecks for attribution,'' in \emph{ICLR}, 2020.

\bibitem{ras2022explainable}
G.~Ras, N.~Xie, M.~Van~Gerven, and D.~Doran, ``Explainable deep learning: A field guide for the uninitiated,'' \emph{Journal of Artificial Intelligence Research}, vol.~73, pp. 329--396, 2022.

\bibitem{kim2017interpretable}
J.~Kim and J.~Canny, ``Interpretable learning for self-driving cars by visualizing causal attention,'' in \emph{ICCV}, 2017, pp. 2942--2950.

\bibitem{mascharka2018transparency}
D.~Mascharka, P.~Tran, R.~Soklaski, and A.~Majumdar, ``Transparency by design: Closing the gap between performance and interpretability in visual reasoning,'' in \emph{CVPR}, 2018, pp. 4942--4950.

\bibitem{zeiler2014visualizing}
M.~D. Zeiler and R.~Fergus, ``Visualizing and understanding convolutional networks,'' in \emph{ECCV}.\hskip 1em plus 0.5em minus 0.4em\relax Springer, 2014, pp. 818--833.

\bibitem{springenberg2014striving}
J.~T. Springenberg, A.~Dosovitskiy, T.~Brox, and M.~Riedmiller, ``Striving for simplicity: The all convolutional net,'' in \emph{ICLR}, 2014.

\bibitem{smilkov2017smoothgrad}
D.~Smilkov, N.~Thorat, B.~Kim, F.~Vi{\'e}gas, and M.~Wattenberg, ``Smoothgrad: removing noise by adding noise,'' in \emph{Workshop, ICML}, 2017.

\bibitem{sundararajan2017axiomatic}
M.~Sundararajan, A.~Taly, and Q.~Yan, ``Axiomatic attribution for deep networks,'' in \emph{ICML}, 2017, pp. 3319--3328.

\bibitem{Naveed}
N.~Akhtar, ``A survey of explainable ai in deep visual modeling: Methods and metrics,'' \emph{arXiv preprint arXiv:2301.13445}, 2023.

\bibitem{omeiza2019smooth}
D.~Omeiza, S.~Speakman, C.~Cintas, and K.~Weldermariam, ``Smooth grad-cam++: An enhanced inference level visualization technique for deep convolutional neural network models,'' \emph{arXiv preprint arXiv:1908.01224}, 2019.

\bibitem{fong2019understanding}
R.~Fong, M.~Patrick, and A.~Vedaldi, ``Understanding deep networks via extremal perturbations and smooth masks,'' in \emph{ICCV}, 2019, pp. 2950--2958.

\bibitem{IGOS}
Q.~Zhongang, K.~Saeed, and F.~Li, ``Visualizing deep networks by optimizing with integrated gradients,'' in \emph{AAAI}, 2020.

\bibitem{fel2023don}
T.~Fel, M.~Ducoffe, D.~Vigouroux, R.~Cad{\`e}ne, M.~Capelle, C.~Nicod{\`e}me, and T.~Serre, ``Don't lie to me! robust and efficient explainability with verified perturbation analysis,'' in \emph{CVPR}, 2023, pp. 16\,153--16\,163.

\bibitem{xie2019visual}
N.~Xie, F.~Lai, D.~Doran, and A.~Kadav, ``Visual entailment: A novel task for fine-grained image understanding,'' \emph{arXiv preprint arXiv:1901.06706}, 2019.

\bibitem{NEURIPS2022_0073cc73}
S.~Stalder, N.~Perraudin, R.~Achanta, F.~Perez-Cruz, and M.~Volpi, ``What you see is what you classify: Black box attributions,'' in \emph{NeurIPS}, vol.~35, 2022, pp. 84--94.

\bibitem{alvarez2018towards}
D.~Alvarez-Melis and T.~S. Jaakkola, ``Towards robust interpretability with self-explaining neural networks,'' \emph{NeurIPS}, 2018.

\bibitem{NEURIPS2019_adf7ee2d}
C.~Chen, O.~Li, D.~Tao, A.~Barnett, C.~Rudin, and J.~K. Su, ``This looks like that: Deep learning for interpretable image recognition,'' in \emph{NeurIPS}, vol.~32, 2019.

\bibitem{koh2020concept}
P.~W. Koh, T.~Nguyen, Y.~S. Tang, S.~Mussmann, E.~Pierson, B.~Kim, and P.~Liang, ``Concept bottleneck models,'' in \emph{ICML}, 2020, pp. 5338--5348.

\bibitem{wang2023botcl}
B.~Wang, L.~Li, Y.~Nakashima, and H.~Nagahara, ``Learning bottleneck concepts in image classification,'' in \emph{CVPR}, 2023.

\bibitem{verma2020counterfactual}
S.~Verma, V.~Boonsanong, M.~Hoang, K.~E. Hines, J.~P. Dickerson, and C.~Shah, ``Counterfactual explanations and algorithmic recourses for machine learning: A review,'' \emph{arXiv preprint arXiv:2010.10596}, 2020.

\bibitem{vandenhende2022making}
S.~Vandenhende, D.~Mahajan, F.~Radenovic, and D.~Ghadiyaram, ``Making heads or tails: Towards semantically consistent visual counterfactuals,'' in \emph{ECCV}, 2022, pp. 261--279.

\bibitem{khorram2022cycle}
S.~Khorram and L.~Fuxin, ``Cycle-consistent counterfactuals by latent transformations,'' in \emph{CVPR}, 2022, pp. 10\,203--10\,212.

\bibitem{goyal2019counterfactual}
Y.~Goyal, Z.~Wu, J.~Ernst, D.~Batra, D.~Parikh, and S.~Lee, ``Counterfactual visual explanations,'' in \emph{ICML}, 2019, pp. 2376--2384.

\bibitem{kommiya2021towards}
R.~Kommiya~Mothilal, D.~Mahajan, C.~Tan, and A.~Sharma, ``Towards unifying feature attribution and counterfactual explanations: Different means to the same end,'' in \emph{AAAI}, 2021, pp. 652--663.

\bibitem{vaswani2017attention}
A.~Vaswani, N.~Shazeer, N.~Parmar, J.~Uszkoreit, L.~Jones, A.~N. Gomez, {\L}.~Kaiser, and I.~Polosukhin, ``Attention is all you need,'' \emph{NeurIPS}, vol.~30, 2017.

\bibitem{devlin2018bert}
J.~Devlin, M.-W. Chang, K.~Lee, and K.~Toutanova, ``Bert: Pre-training of deep bidirectional transformers for language understanding,'' \emph{arXiv preprint arXiv:1810.04805}, 2018.

\bibitem{parmar2018image}
N.~Parmar, A.~Vaswani, J.~Uszkoreit, L.~Kaiser, N.~Shazeer, A.~Ku, and D.~Tran, ``Image transformer,'' in \emph{ICML}, 2018, pp. 4055--4064.

\bibitem{carion2020end}
N.~Carion, F.~Massa, G.~Synnaeve, N.~Usunier, A.~Kirillov, and S.~Zagoruyko, ``End-to-end object detection with transformers,'' in \emph{ECCV}.\hskip 1em plus 0.5em minus 0.4em\relax Springer, 2020, pp. 213--229.

\bibitem{dosovitskiy2020image}
A.~Dosovitskiy, L.~Beyer, A.~Kolesnikov, D.~Weissenborn, X.~Zhai, T.~Unterthiner, M.~Dehghani, M.~Minderer, G.~Heigold, S.~Gelly \emph{et~al.}, ``An image is worth 16x16 words: Transformers for image recognition at scale,'' in \emph{ICLR}, 2021.

\bibitem{wen2022self}
X.~Wen, B.~Zhao, A.~Zheng, X.~Zhang, and X.~Qi, ``Self-supervised visual representation learning with semantic grouping,'' \emph{NeurIPS}, vol.~35, pp. 16\,423--16\,438, 2022.

\bibitem{jiang2023object}
J.~Jiang, F.~Deng, G.~Singh, and S.~Ahn, ``Object-centric slot diffusion,'' \emph{arXiv preprint arXiv:2303.10834}, 2023.

\bibitem{xu2022groupvit}
J.~Xu, S.~De~Mello, S.~Liu, W.~Byeon, T.~Breuel, J.~Kautz, and X.~Wang, ``Groupvit: Semantic segmentation emerges from text supervision,'' in \emph{CVPR}, 2022, pp. 18\,134--18\,144.

\bibitem{deng2009imagenet}
J.~Deng, W.~Dong, R.~Socher, L.-J. Li, K.~Li, and L.~Fei-Fei, ``Imagenet: A large-scale hierarchical image database,'' in \emph{CVPR}, 2009, pp. 248--255.

\bibitem{caltech}
G.~Griffin, A.~Holub, and P.~Perona, ``{Caltech256 object category dataset},'' California Institute of Technology, Tech. Rep., 2007.

\bibitem{ACRIMA}
A.~Diaz-Pinto, S.~Morales, V.~Naranjo, T.~K{\"o}hler, J.~M. Mossi, and A.~Navea, ``{CNNs} for automatic glaucoma assessment using fundus images: an extensive validation,'' \emph{Biomedical Engineering Online}, vol.~18, no.~1, p.~29, 2019.

\bibitem{wang2017chestx}
X.~Wang, Y.~Peng, L.~Lu, Z.~Lu, M.~Bagheri, and R.~M. Summers, ``Chestx-ray8: Hospital-scale chest x-ray database and benchmarks on weakly-supervised classification and localization of common thorax diseases,'' in \emph{CVPR}, 2017, pp. 2097--2106.

\bibitem{loshchilov2017decoupled}
I.~Loshchilov and F.~Hutter, ``Decoupled weight decay regularization,'' \emph{arXiv preprint arXiv:1711.05101}, 2017.

\bibitem{rw2019timm}
R.~Wightman, ``Pytorch image models,'' 2019.

\bibitem{Wu_Palmer}
Z.~Wu and M.~Palmer, ``Verbs semantics and lexical selection,'' in \emph{ACL}, 1994, p. 133–138.

\bibitem{miller1995wordnet}
G.~A. Miller, ``Wordnet: a lexical database for english,'' \emph{Communications of the ACM}, vol.~38, no.~11, pp. 39--41, 1995.

\bibitem{radford2021learning}
A.~Radford, J.~W. Kim, C.~Hallacy, A.~Ramesh, G.~Goh, S.~Agarwal, G.~Sastry, A.~Askell, P.~Mishkin, J.~Clark \emph{et~al.}, ``Learning transferable visual models from natural language supervision,'' in \emph{ICML}, 2021, pp. 8748--8763.

\bibitem{top_down}
J.~Zhang, S.~A. Bargal, Z.~Lin, J.~Brandt, X.~Shen, and S.~Sclaroff, ``Top-down neural attention by excitation backprop,'' \emph{IJCV}, vol. 126, no.~10, pp. 1084--1102, 2018.

\bibitem{yeh2019fidelity}
C.-K. Yeh, C.-Y. Hsieh, A.~Suggala, D.~I. Inouye, and P.~K. Ravikumar, ``On the (in) fidelity and sensitivity of explanations,'' \emph{NeurIPS}, vol.~32, 2019.

\bibitem{he2016deep}
K.~He, X.~Zhang, S.~Ren, and J.~Sun, ``Deep residual learning for image recognition,'' in \emph{CVPR}, 2016, pp. 770--778.

\bibitem{zhang2022resnest}
H.~Zhang, C.~Wu, Z.~Zhang, Y.~Zhu, H.~Lin, Z.~Zhang, Y.~Sun, T.~He, J.~Mueller, R.~Manmatha \emph{et~al.}, ``Resnest: Split-attention networks,'' in \emph{CVPR}, 2022, pp. 2736--2746.

\bibitem{huang2017densely}
G.~Huang, Z.~Liu, L.~Van Der~Maaten, and K.~Q. Weinberger, ``Densely connected convolutional networks,'' in \emph{CVPR}, 2017, pp. 4700--4708.

\bibitem{howard2017mobilenets}
A.~G. Howard, M.~Zhu, B.~Chen, D.~Kalenichenko, W.~Wang, T.~Weyand, M.~Andreetto, and H.~Adam, ``Mobilenets: Efficient convolutional neural networks for mobile vision applications,'' \emph{arXiv preprint arXiv:1704.04861}, 2017.

\bibitem{tan2019efficientnet}
M.~Tan and Q.~Le, ``Efficientnet: Rethinking model scaling for convolutional neural networks,'' in \emph{ICML}, 2019, pp. 6105--6114.

\bibitem{szegedy2016rethinking}
C.~Szegedy, V.~Vanhoucke, S.~Ioffe, J.~Shlens, and Z.~Wojna, ``Rethinking the inception architecture for computer vision,'' in \emph{CVPR}, 2016, pp. 2818--2826.

\bibitem{szegedy2017inception}
C.~Szegedy, S.~Ioffe, V.~Vanhoucke, and A.~Alemi, ``Inception-v4, inception-resnet and the impact of residual connections on learning,'' in \emph{AAAI}, vol.~31, no.~1, 2017.

\bibitem{zhang2021group}
Q.~Zhang, L.~Rao, and Y.~Yang, ``Group-cam: Group score-weighted visual explanations for deep convolutional networks,'' \emph{arXiv preprint arXiv:2103.13859}, 2021.

\bibitem{tursun2022sess}
O.~Tursun, S.~Denman, S.~Sridharan, and C.~Fookes, ``Sess: Saliency enhancing with scaling and sliding,'' in \emph{ECCV}.\hskip 1em plus 0.5em minus 0.4em\relax Springer, 2022, pp. 318--333.

\bibitem{khorram2021igos++}
S.~Khorram, T.~Lawson, and L.~Fuxin, ``{iGOS}++ integrated gradient optimized saliency by bilateral perturbations,'' in \emph{Conference on Health, Inference, and Learning}, 2021, pp. 174--182.

\bibitem{GreedyAS}
Cheng-Yu, C.-K. Hsieh, X.~Yeh, P.~Liu, S.~Ravikumar, S.~Kim, Kuma, and C.-J. Hsieh, ``Evaluations and methods for explanation through robustness analysis,'' in \emph{ICLR}, 2021.

\bibitem{orlando2020refuge}
J.~I. Orlando, H.~Fu, J.~B. Breda, K.~Van~Keer, D.~R. Bathula, A.~Diaz-Pinto, R.~Fang, P.-A. Heng, J.~Kim, J.~Lee \emph{et~al.}, ``Refuge challenge: A unified framework for evaluating automated methods for glaucoma assessment from fundus photographs,'' \emph{Medical image analysis}, vol.~59, p. 101570, 2020.

\end{thebibliography}

%
\begin{IEEEbiography}[{\includegraphics[width=1in,height=1.25in,clip,keepaspectratio]{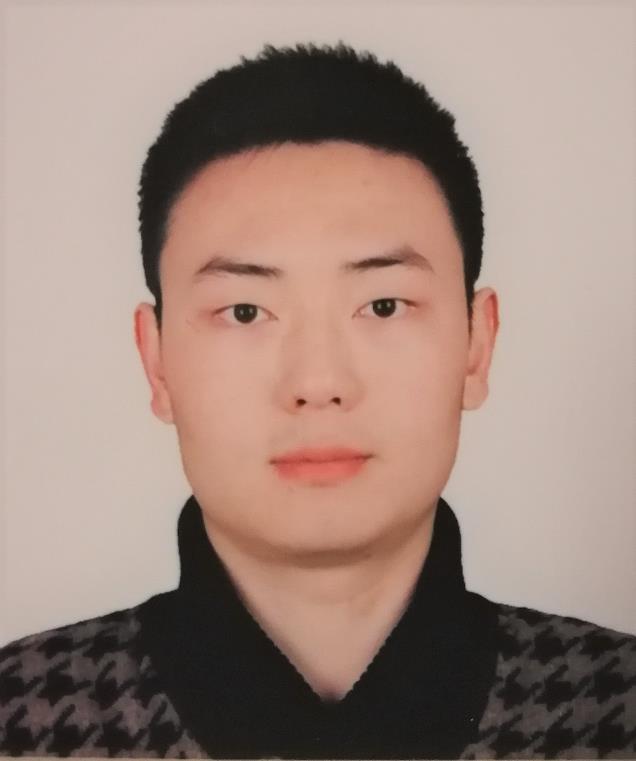}}]{Bowen Wang} received a B.S. and M.S. degree in Computer Science from Anhui University, China, in 2016 and in Medical Information from Osaka University, Japan, in 2020, respectively, and a Ph.D. degree in Computer Science from Osaka University, in 2023. He has received the best paper award in APAMI 2020 and is a member of IEEE. He is currently a specially-appointed researcher with the Institute for Datability Science (IDS), Osaka University, Japan.  His research interests in computer vision, explainable AI, city perception, and medical AI.
\end{IEEEbiography}

\begin{IEEEbiography}[{\includegraphics[width=1in,height=1.25in,clip,keepaspectratio]{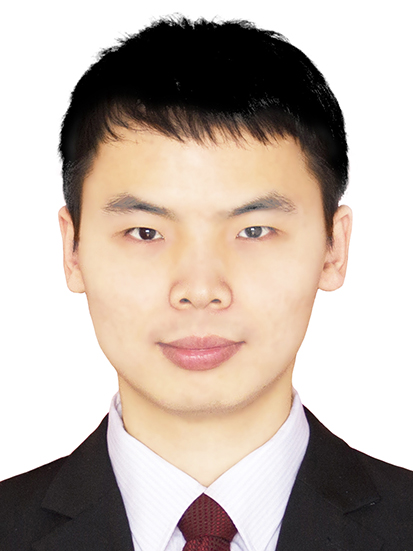}}]{Liangzhi Li} currently serves as the Chief Scientist at Xiamen Meet You Co., Ltd, and as the Head of Meetyou AI Lab (MAIL) since 2023. He also holds a professorship at the Computer Science Department of Qufu Normal University from the same year. He received the B.S and M.S degrees in Computer Science from South China University of Technology (SCUT), China, in 2012 and 2016, respectively, and the Ph.D. degree in Engineering from Muroran Institute of Technology, Japan, in 2019. After his graduation, he worked as a researcher (2019-2021) at Osaka University and later served as an Assistant Professor (2021-2023). His main fields of research interest include computer vision, explainable AI, and medical imaging. In his current roles, he is focusing on advancing technology-driven solutions for business enhancements, contributing to the field of computer science through educational excellence and carrying out pioneering research in the realm of artificial intelligence.
\end{IEEEbiography}

\begin{IEEEbiography}[{\includegraphics[width=1in,height=1.25in,clip,keepaspectratio]{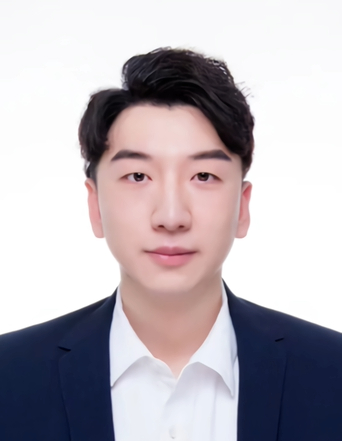}}]{Jiahao Zhang} earned his Bachelor's degree in Computer Science from Zhengzhou University, China, and his Master's degree in Medical Information Research from Osaka University. He is currently pursuing his Ph.D. at the Institute for Datability Science, Osaka University. His research focuses on computer vision and medical AI.
\end{IEEEbiography}

\begin{IEEEbiography}[{\includegraphics[width=1in,height=1.25in,clip,keepaspectratio]{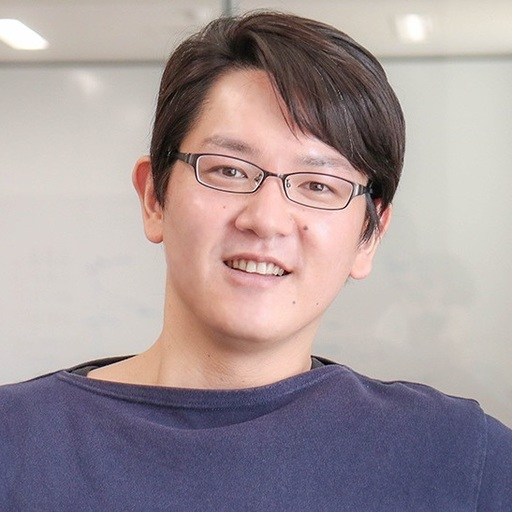}}]{Yuta Nakashima} (M’09) received the B.E. and M.E. degrees in communication engineering and the Ph.D. degree in engineering from Osaka University, Osaka, Japan, in 2006, 2008, and 2012, respectively. From 2012 to 2016, he was an Assistant Professor at the Nara Institute of Science and Technology. He is currently a Professor at the Institute for Datability Science, Osaka University. He was a Visiting Scholar at the University of North Carolina at Charlotte in 2012 and at Carnegie Mellon University from 2015 to 2016. His research interests include computer vision and machine learning and their applications. His main research includes video content analysis using machine learning approaches. Prof. Nakashima is a member of ACM, IEICE, and IPSJ.
\end{IEEEbiography}

\begin{IEEEbiography}[{\includegraphics[width=1in,height=1.25in,clip,keepaspectratio]{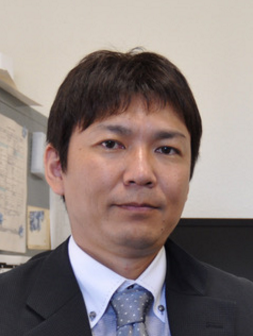}}]{Hajime Nagahara} received the Ph.D. degree in system engineering from Osaka University, Suita, Japan, in 2001. Since 2017, he is a Professor with the Institute for Datability Science, Osaka University. He was a Research Associate with the Japan Society for the Promotion of Science from 2001 to 2003. He was an Assistant Professor with the Graduate School of Engineering Science, Osaka University, from 2003 to 2010. He was an Associate Professor with the Faculty of Information Science and Electrical Engineering, Kyushu University, from 2010 to 2017. He was a Visiting Associate Professor with CREA University of Picardie Jules Verns, in 2005. He was a Visiting Researcher with Columbia University in 2007–2008 and 2016–2017. His research interests include computational photography and computer vision. He was the recipient of an ACM VRST2003 Honorable Mention Award in 2003, IPSJ Nagao Special Researcher Award in 2012, ICCP2016 Best Paper Runners-up, and SSII Takagi Award in 2016.
\end{IEEEbiography}

\vfill

\end{document}